\documentclass[journal]{IEEEtran}
\usepackage{times}
\usepackage[numbers]{natbib}
\usepackage{multicol}
\usepackage{graphicx}
\usepackage[cmex10]{amsmath}
\usepackage[caption=false]{subfig}
\usepackage{fixltx2e}
\usepackage[ruled, vlined]{algorithm2e}

\begin{document}

\title{Qualitative Relational Mapping and Navigation for Planetary Rovers}

\author{
  Mark McClelland%
    \thanks{Post-Doctoral Associate, Cornell University, mjm496@cornell.edu.} and Mark Campbell\thanks{Professor, Director of the Sibley School of Mechanical and Aerospace Engineering, Cornell University, IEEE Member, mc288@cornell.edu.}\\
  {\normalsize\itshape
   Dept. of Mechanical and Aerospace Engineering, Cornell University, Ithaca, NY, 14853, USA}\\
  %\and
  Tara Estlin%
   \thanks{Deputy Technologist, Mission Systems and Operations Division, AIAA Member, Tara.Estlin@jpl.nasa.gov.}\\
  {\normalsize\itshape
  Jet Propulsion Laboratory, California Institute of Technology, Pasadena, CA, 91109}
 }

\maketitle

\begin{abstract}
This paper presents a novel method for qualitative mapping of large scale spaces. The proposed framework makes use of a graphical representation of the world in order to build a map consisting of qualitative constraints on the geometric relationships between landmark triplets. A novel measurement method based on camera imagery is presented which extends previous work from the field of Qualitative Spatial Reasoning. Measurements are fused into the map using a deterministic, iterative graph update. A Branch-and-Bound approach is taken to solve a set of non-convex feasibility problems required for generating on-line measurements and off-line operator lookup tables. A navigation approach for travel between distant landmarks is developed, using estimates of the Relative Neighborhood Graph extracted from the qualitative map in order to generate a sequence of landmark objectives based on proximity. Average and asymptotic performance of the mapping algorithm is evaluated using Monte Carlo tests on randomly generated maps, and data-driven simulation results are presented for a robot traversing the Jet Propulsion Laboratory Mars Yard while building a relational map. Simulation results demonstrate an initial rapid convergence of qualitative state estimates for visible landmarks, followed by a slow tapering as the remaining ambiguous states are removed from the map.
\end{abstract}

\IEEEpeerreviewmaketitle

\section{Introduction}
\label{sec:Intro}
When available, absolute position sensors such as GPS systems provide high quality measurements for generating the position and heading estimates necessary for long-distance autonomous robotic operation. Unfortunately, such systems are unavailable for a number of applications, including extra-planetary exploration, operation in GPS-denied regions, and operation of extremely small or low-cost robotic platforms. 

In the absence of absolute position sensors, existing robot localization systems tend to either rely solely on local sensors of ego-motion (such as Inertial-Measurement Units and wheel encoders) as in the current GESTALT system for the Mars Exploration Rovers (MER) discussed by \citet{Ali05}, or incorporate measurements of the rover's relative position and orientation with respect to certain landmarks in the environment using vision or ranging sensors. This may consist of triangulation from known reference positions as demonstrated by \citet{Kuipers88}, or the construction of adaptive feature maps as in the Simultaneous Localization and Mapping (SLAM) framework \cite{Thrun06}. These methods have definite strengths, including the ability to provide both global position and orientation estimates as well as accurate estimates of the uncertainty in the parameters. They can also provide  localization of environmental features in the global reference frame and thus allow the accumulation of information for the assembly of the stable maps necessary for long-distance planning. However, these approaches often face a number of limitations, including computational expense, a reliance on point estimates of landmarks, and the need for high quality sensing to determine metrical distance measurements to visible landmarks. In contrast, the motivation behind this work is to extract information about objects of interest from a minimal set of low-cost sensors, in this case a single camera without any estimates of ego-motion.

\begin{figure*}[!t]
\centering
\includegraphics[width=7in]{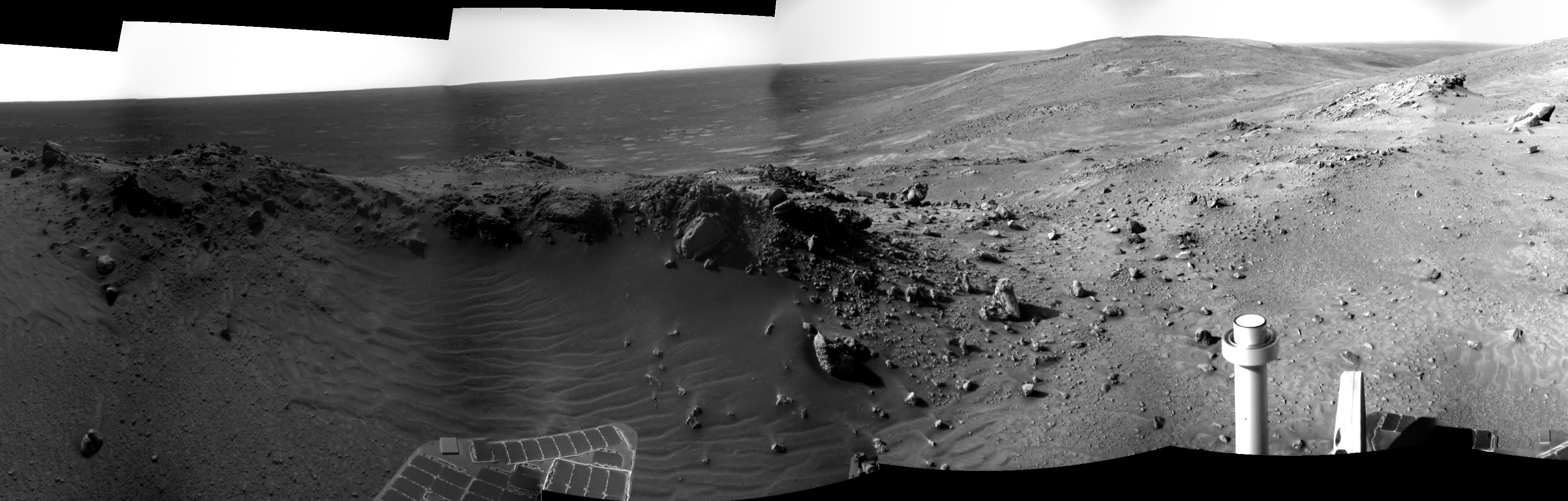}
\caption{Example of a Martian landscape, taken by the Spirit rover on Sol 476. The area shown in this figure is characterized by large open spaces with scattered landmarks.}
\label{fig:mars}
\end{figure*}

The solution to the problem of long-term autonomy in the absence of global reference data proposed in this paper is process called the `Qualitative Relational Map' (QRM), in which the relative geometrical relationships between landmarks are tracked using qualitative information inferred from camera images. The key novelties in this work are a representation of geometrical relationships that defines both qualitative orientations and distances, a method for extracting and fusing measurements of qualitative states using global nonlinear optimization, and the implementation of the mapping system in a realistic experimental scenario with data gathered in the JPL Mars Yard. The test case used to evaluate system performance is the exploration and mapping of a Mars-like environment; this application is characterized by large open areas with clusters of interesting features such as the region shown in Figure \ref{fig:mars}. The QRM system developed in this work is an extension of the online mapping process presented in \cite{McClelland10}.

A key aim of the proposed qualitative framework is to decouple the robot position estimation problem from that of map building. This is inspired by the insight that many robot tasks, such as navigation, do not require a fully defined metrical map. Use of qualitative relations between landmarks allows maps to remain useful in the presence of the distortion that may occur in traditional metrical mapping approaches due to wheel slippage, rate gyro biases, etc. These sensing errors lead to uncertain estimates of robot ego-motion, which can induce filter inconsistencies in traditional metrical SLAM, as observed by \citet{Julier01} and \citet{Huang07}, particularly when using the bearing only measurements provided by monocular cameras. SLAM inconsistencies have been addressed in a number of ways in the literature; such as the ego-frame approach with linked submaps presented by \citet{Castellanos07}, the topological methods presented by \citet{Randell92} and\citet{Angeli08}, the topometric mapping discussed by \citet{Sibley10}, and the place-base mapping discussed by \citet{Cummins08b}. These approaches are often successful at limiting filter inconsistencies and map drift in indoors or in urban environments, however, they face a number of challenges in large, unstructured environments. Such areas lack the high feature densities necessary for generating well-defined places or submaps, and do not have the limited connectivities between areas required for topological reasoning.

The qualitative approach detailed in this paper avoids the consistency problem entirely by extracting geometrical constraints on landmark relationships directly from camera measurements, rather than relying on estimated ego-motion. Navigation objectives can then be expressed in terms of these relationships. For example, `stay to the right of points A and B' can be re-expressed in terms of a sequence of desired qualitative states with respect to the map graph. Representing landmark relationships qualitatively avoids both integration and linearization errors, but does so at the cost of maintaining scale free maps with large uncertainties in exact landmark positions, particularly at the edge of the map. In essence, this can be seen as a trade-off between map precision and map consistency. Just as there are an infinite number of spatial layouts that may satisfy any given topological specification, there are an infinite number of metrical arrangements of landmarks that have equivalent qualitative maps. However, the coordinate sets for all of these point arrangements are constrained to satisfy a set nonlinear inequalities implied by the qualitative statements in the map. Thus, one interpretation of the QRM algorithm is as a form of topological-style reasoning operating on topologically ambiguous spaces.

One approach of how to represent the `shape' of a set of points has been that of statistical shape theory, which defines `shape' to be what remains once scale, rotation, and translation effects have been removed via dimensional reduction. The approach discussed by \citet{Dryden93} and \citet{Mitteroecker09} uses a QR decomposition to transform a set of high dimensional points to the surface of a hypersphere in a scale, rotation, and translation invariant subspace. Continuous deformation of point sets correspond to trajectories over the hypersphere, and a statistical similarity metric can be constructed by considering probability distributions over the hypersphere. The relationships encoded in the proposed QRM, although driven by different geometrical concerns, correspond to defining nonlinear constraints on these point distributions. Landmark arrangements that have the same qualitative will occupy a bounded, though non-convex, region of the hypersphere defined by the inequality constraints which correspond to the qualitative states encoded in the map edges. Critically, while statistical shape theory requires access to the true landmark locations in some reference frame in order to calculate the `shape' of a point set, the QRM learns the constraints without attempting to estimate the locations themselves.

Previous work on qualitative mapping and navigation for ground robots includes the QUALNAV system described by \citet{Levitt90}, which relied on binary relationships inferred from the cyclical ordering of landmarks in a robots view. This representation decomposed the space around landmarks into regions defined by the rays passing through each landmark pair, as crossing those lines swaps landmark position in the view. Cyclical ordering was also used by \citet{Wallgrun10} to learn the topologies of environments made up of hallway junctions, where junctions are labeled according to their qualitative cardinal orientation. The representation was extended by \citet{Schlieder93} to include the directions opposite landmarks in order to eliminate map ambiguities and termed the `panorama'. The cyclical order is also revised to include extended objects with occlusions by \citet{Fogliaroni09}, in which a topological map of visibility regions is induced by tangent lines from the extrema of convex polygonal obstacles. These regions are then learned either from an known map of object shapes and locations, or by an exhaustive search of the space.

The remainder of this paper is laid out as follows. Section \ref{sec:Geometry} contains a background on the formalism used to define qualitative spatial relationships used in this work. Section \ref{sec:Operators} discusses the generation of lookup tables for operators used to manipulate qualitative relationships. Section \ref{sec:Measurements} presents a novel method for generating measurements of qualitative states using camera images. Section \ref{sec:Feasibility} presents a Branch-and-Bound algorithm for solving the non-convex quadratic feasibility problems required to generate measurements and operator tables. Section \ref{sec:Mapping} defines the map structure and summarizes the measurement update algorithm. Section \ref{sec:Navigation} presents a method for extracting estimates of the Relative Neighborhood Graph from qualitative maps, as well as a long-distance navigation strategy based on Voronoi regions. Section \ref{sec:Results} presents the results of a set of Monte-Carlo tests used to evaluate average and asymptotic performance of the mapping algorithm as a function of the number of landmarks simultaneously observed. The results of a data-driven simulation are also presented for a robot moving through a Mars-like environment.
 
\section{Qualitative Relational Geometry}
\label{sec:Geometry}
\begin{figure}[t]
\centering
\subfloat[][Region Boundaries]{\includegraphics[width=1.75in]{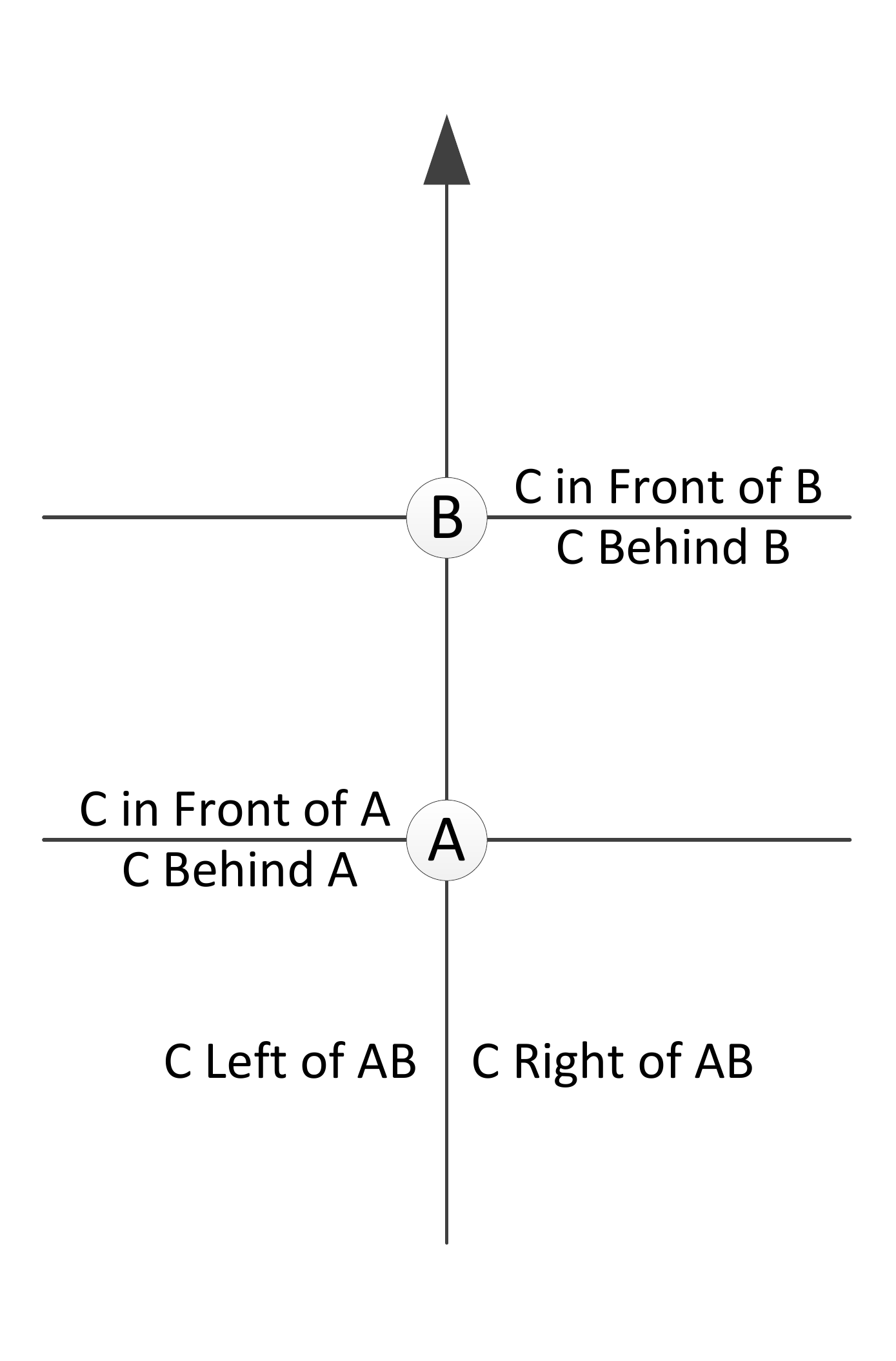} \label{fig:FDCLines}}
\subfloat[][Qualitative Regions.]{\includegraphics[width=1.75in]{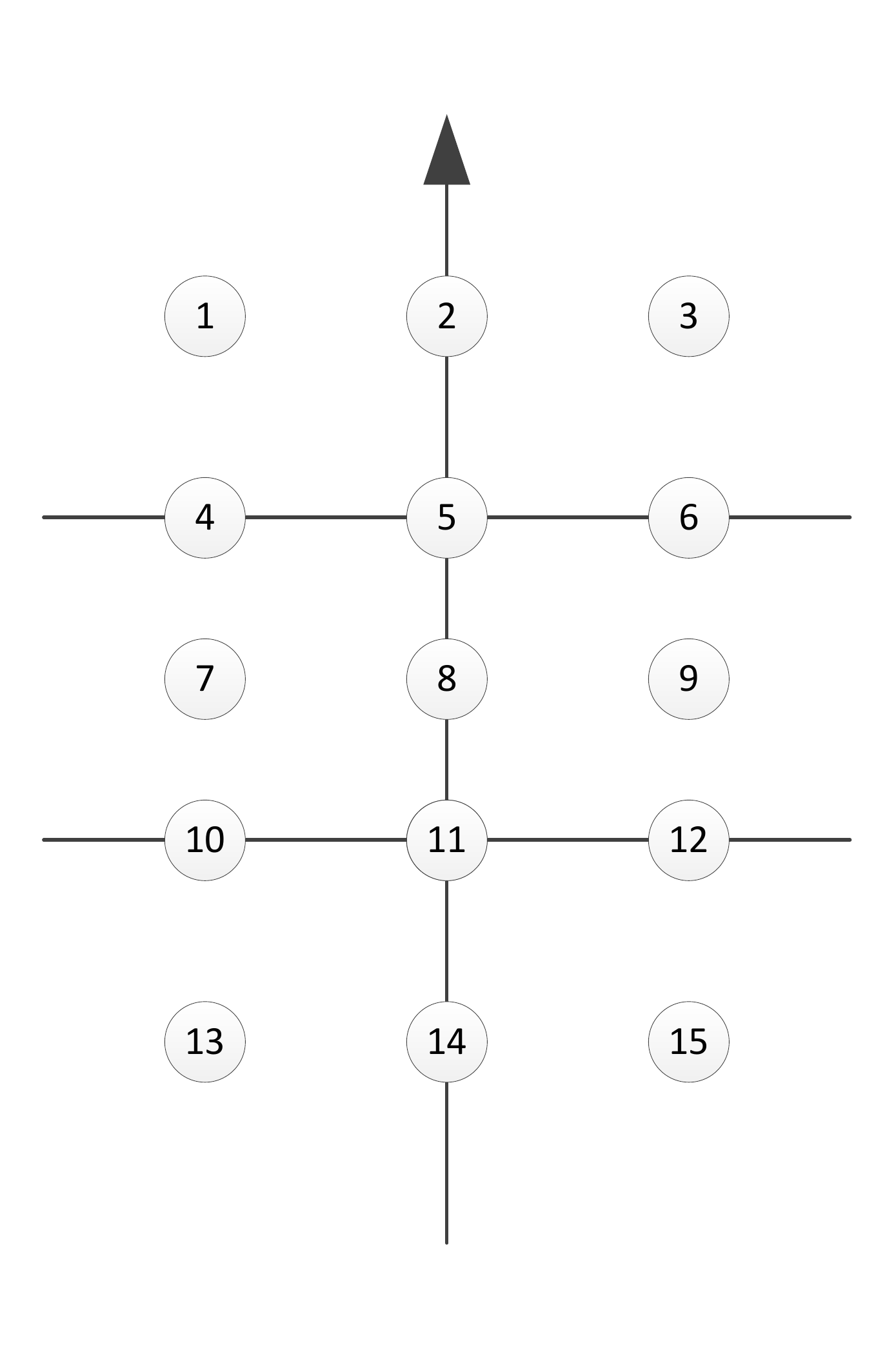} \label{fig:FDCRegions}}
\caption{Schematics of Freksa's Double Cross for 2 landmarks $A$ and $B$. (a) shows the three dichotomies which split up the space around the vector AB. (b) shows the 15 qualitative regions in which a third landmark $C$ can lie that result from these dichotomies.}
\label{fig:FDC}
\end{figure}

Qualitative statements of geometrical relationships require that the 2D plane around landmarks be segmented into discrete regions. The approach presented in this paper makes use of an extension of the double cross discretization based on triplets of landmarks proposed by \citet{Freksa92}. Freksa's Double Cross (FDC) specifies the position of a query point $C$ to $\overline{AB}$, the vector from point $A$ to point $B$, by stating that $C$ can be either to the left or right of $\overline{AB}$, in front or behind $A$ relative to the direction of $\overline{AB}$, and in front or behind $B$ relative to the direction of $\overline{AB}$. These three statements are equivalent to defining the separating boundaries shown in Figure \ref{fig:FDCLines}. If the boundary lines are also included as states, this results in the 15 possible geometrical relationships between $C$ and $\overline{AB}$ shown in Figure \ref{fig:FDCRegions}. 

\begin{figure}[tbh!]
\centering
\subfloat[][Region Boundaries]{\includegraphics[width=3in]{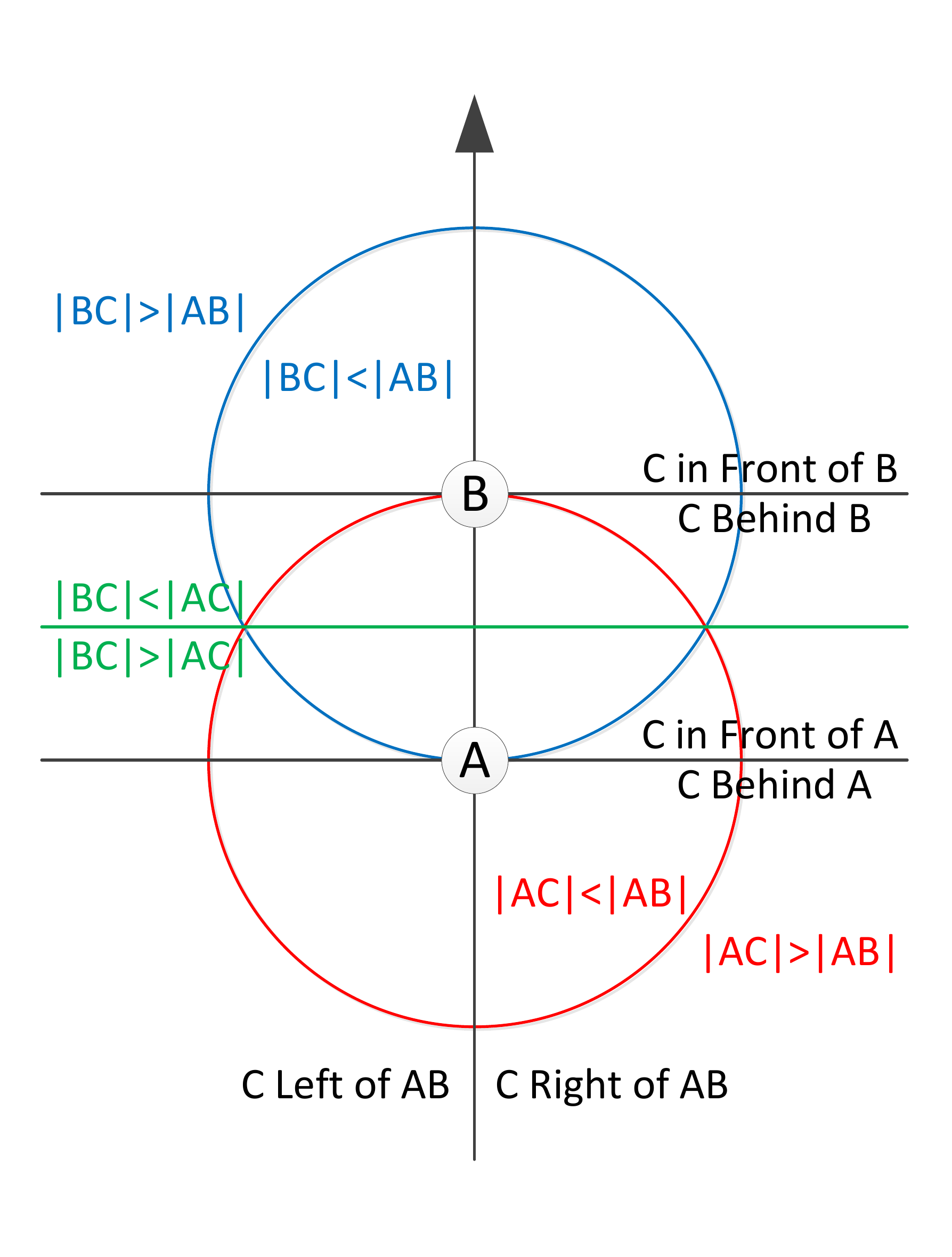} \label{fig:EDCLines}}\\
\subfloat[][Qualitative Regions.]{\includegraphics[width=3in]{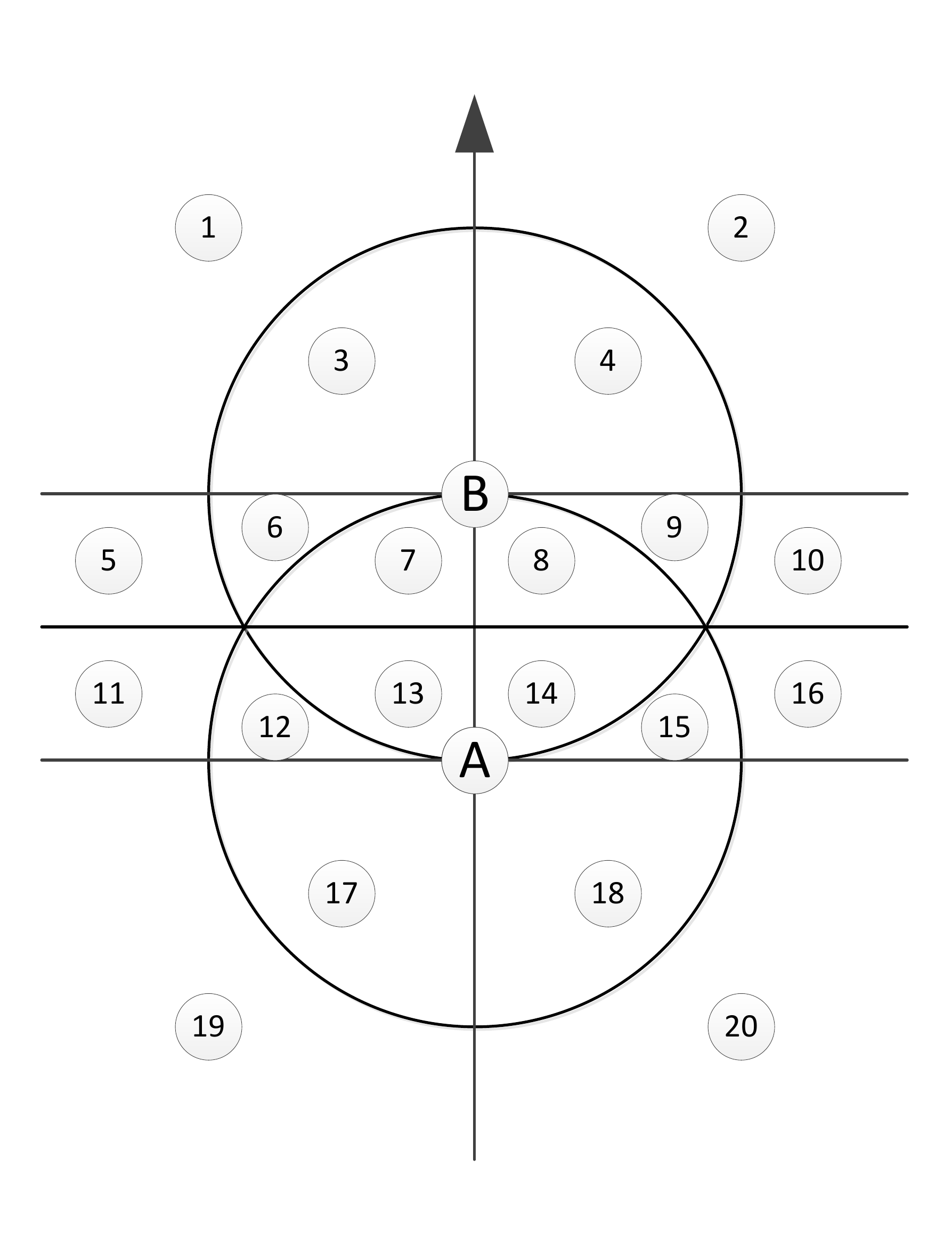} \label{fig:EDCRegions}}
\caption{Schematics of the Extended Double Cross (EDC) for two landmarks $A$ and $B$. (a) six dichotomies splitting the space around the vector AB. (b) 20 qualitative regions where a third landmark $C$ can lie that result from the dichotomies in (a).}
\label{fig:EDC}
\end{figure}

The work in this paper defines an Extended Double Cross (EDC), which adds the additional statements that compare the distance from $C$ to $A$ against that from $C$ to $B$, the distance from $C$ to $A$ with that between $A$ and $B$, and the distance from $C$ to $B$ with that between $A$ and $B$. The separating boundaries associated with the EDC representation are shown in Figure \ref{fig:EDCLines}, and the $20$ possible regions between boundaries are labeled in Figure \ref{fig:EDCRegions}. The FDC representation can be interpreted as qualitatively specifying the angles in the triangle ABC, while the EDC adds explicit qualitative statements about the edge lengths $|AB|$, $|BC|$, and $|CA|$.

The relationship between the point $C$ and $\overline{AB}$ is denoted as the `qualitative state' $AB:C$, which can be one of 20 regions, as shown in Figure \ref{fig:EDCRegions}. In general, there may be insufficient information available to determine exactly which EDC region around $\overline{AB}$ contains the point $C$, in which case the state $AB:C$ will indicate a set of possible EDC regions. For the sake of clarity, the qualitative representation used in this paper is restricted to considering only the relationships defined by the 20 regions defined by the separating lines shown in Figure \ref{fig:EDCLines}, which can be expressed in terms of linear and quadratic inequalities. In most practical implementations, this is sufficient because physical landmarks are unlikely to lie exactly on a boundary line. If necessary, the optimization approach detailed in the following sections can be easily extended to equality constraints in order to include the lines and line intersections as additional states, or the lines may be incorporated into neighboring regions.

\section{EDC Operators}
\label{sec:Operators}
Building a cohesive map of landmark relationships from disparate camera measurements requires the ability to infer how observed relationships can restrict the states of unobserved ones. Doing so requires three unary operators that convert between different representations of a given landmark triple and a composition operator that uses known information about two qualitative states in order to reason about a third. These operators enable transformations between EDC states in the same manner as the FDC operators discussed by \citet{Scivos05}.

\subsection{Unary Operators}
The qualitative relationships between points $A$, $B$, and $C$ can be stored in the EDC states for any of $\{AB:C,BC:A,CA:B,BA:C,AC:B,CB:A\}$. These states are highly interdependent; conversion between the triples is straightforward using two cyclical permutation operations to generate $BC:A$ and $CA:B$ given $AB:C$, and an inversion operation to determines $BA:C$ given $AB:C$. The left shift operator is denoted as $\text{LEFT}(AB:C) = BC:A$, while the right shift operator is denoted as $\text{RIGHT}(AB:C) = CA:B$, and the inverse operator is denoted as $\text{INVERSE}(AB:C) = BA:C$. The results of these unary operations can be readily found by inspection of the relevant geometry and are listed in Table \ref{table:Unitary}. Unfortunately, while the inversion operator provides a one-to-one mapping, there are four states which are ambiguous under the cyclical transforms. The ambiguities introduced by these operators are similar to those discussed by \citet{Scivos01}.

\begin{table}[tbh]
\centering
\caption{EDC Unary Transformations}
\label{table:Unitary}
\begin{tabular}{c|c|c|c}
$AB:C$ & $BC:A$ & $CA:B$ & $BA:C$ \\
\hline\hline
& & & \\
1 & 17 & 7 & 20\\
2 & 18 & 8 & 19\\
3 & 19 & 13 & 18\\
4 & 20 & 14 & 17\\
5 & 12 & 7 & 16\\
6 & 11 & 13 & 15\\
7 & \{1,5\} & \{12,17\} & 14\\
8 & \{2,10\} & \{15,18\} & 13\\
9 & 16 & 14 & 12\\
10 & 15 & 8 & 11\\
11 & 13 & 6 & 10\\
12 & 7 & 5 & 9\\
13 & \{3,6\} & \{11,19\} & 8\\
14 & \{4,9\} & \{16,20\} & 7\\
15 & 8 & 10 & 6\\
16 & 14 & 9 & 5\\
17 & 7 & 1 & 4\\
18 & 8 & 2 & 3\\
19 & 13 & 3 & 2\\
20 & 14 & 4 & 1
\end{tabular}
\end{table}

\subsection{Composition Operator}
The composition operator determines which EDC states for $AB:D$ are consistent given observed states for $AB:C$ and $BC:D$. While determining the composition rule for any given pair of EDC states for $AB:C$ and $BC:D$ by inspection is a straightforward process, the number of combinations required to fully populate the operator table renders accurate manual calculation impractical. Instead, the problem can be formulated as determining the feasibility of a set of inequality constraints that can be automatically defined and solved offline for each state combination. Let the points $A,B,C,D$ be generally defined as $A=(0,0)$, $B=(0,1)$, $C=(\alpha, \beta)$, and $D=(\gamma, \delta)$. Specifying a state for $AB:C$ is equivalent to defining a set of inequalities drawn from the upper third of Table \ref{table:Composition} that point $C$ must satisfy. For example, $AB:C=2$ is equivalent to requiring that
\begin{align*}
 \alpha & > 0\\
 \beta - 1 & > 0\\
 \alpha^2+\beta^2-2\beta& > 0
\end{align*}
Similarly, the EDC states for $BC:D$ are equivalent to inequality sets drawn from the middle block of Table \ref{table:Composition}, while those for $AB:D$ are drawn from the lower third of Table \ref{table:Composition}. The problem of determining if the composition table entry for a pair of states $AB:C$ and $BC:D$ should include a given state for $AB:D$ is accomplished by searching for a point $(\alpha, \beta, \gamma, \delta)$ that jointly satisfies the associated inequality constraints. An efficient Branch-and-Bound algorithm for solving these problems offline is detailed in Section \ref{sec:Feasibility}. An examination of the EDC geometry indicates that any feasible region for this problem will both occupy a non-zero volume of the search space and extend close to the origin. Thus, it is reasonable to also include upper and lower bounds on $(\alpha, \beta, \gamma, \delta)$, so long as those bounds are large compared to $|AB|=1$.

\begin{table}[!ht]
\centering
\caption{EDC boundary expressions for $A=(0,0)$, $B=(0,1)$, $C=(\alpha,\beta)$, $D=(\gamma,\delta)$}
\label{table:Composition}
\begin{tabular}{c|c}
Expression & Interpretation of Expression $<0$\\
\hline\hline
 & \\
$-\alpha$ & $C$ is to the right of $\overline{AB}$\\
$-\beta$ & $C$ is in front of $A$ wrt $\overline{AB}$\\
$1-\beta$ & $C$ is in front of $B$ wrt $\overline{AB}$\\
$1 - 2\beta$ & $|AC| > |BC|$\\
$1-(\alpha^2+\beta^2)$ & $|AC| > |AB|$\\
$2\beta-(\alpha^2+\beta^2)$ & $|BC| > |AB|$\\
& \\
\hline
& \\
$(\alpha \delta + \gamma)-(\alpha + \beta \gamma)$ & $D$ is to the right of $\overline{BC}$\\
$(\beta + \delta)-(\beta \delta + \alpha \gamma + 1)$ & $D$ is in front of $B$ wrt $\overline{BC}$\\
$(\alpha^2+\beta^2+\delta)-(\beta \delta + \alpha \gamma + \beta)$ & $D$ is in front of $C$ wrt $\overline{BC}$\\
$(\alpha^2+\beta^2+2\delta)-(2\beta \delta + 2\alpha \gamma + 1)$ & $|BD| > |CD|$\\
$(\alpha^2+\beta^2+2\delta)-(\gamma^2+\delta^2+2\beta)$ & $|BD| > |BC|$\\
$(2\alpha \gamma + 2\beta \delta + 1)-(\gamma^2+\delta^2+2\beta)$ & $|CD| > |BC|$\\
& \\
\hline
& \\
$-\gamma$ & $D$ is to the right of $\overline{AB}$\\
$-\delta$ & $D$ is in front of $A$ wrt $\overline{AB}$\\
$1-\delta$ & $D$ is in front of $B$ wrt $\overline{AB}$\\
$1-2\delta$ & $|AD| > |BD|$\\
$1-(\gamma^2+\delta^2)$ & $|AD| > |AB|$\\
$2\delta-(\gamma^2+\delta^2)$ & $|BD| > |AB|$\\
\end{tabular}
\end{table}

\subsection{Operators Example}
The use of these operators on EDC states is best illustrated by a simple example. Consider the case of four landmarks, $A$, $B$, $C$, and $D$. Let $X$ represent the set of qualitative states $AB:C = \{6,7\}$, $Y$ the state $AC:D = \{16\}$, and $Z$ the states $BC:D = \{1,5,11,12,17,18,19,20\}$. The EDC operators can be used to show that $X$ and $Y$ imply $Z$, or more specifically that $Z= \text{COMPOSE}(\text{LEFT}(X), \text{INVERSE}(Y))$. Evaluation of this expression is done as follows. Performing a left shift on $X$ is done by finding the mappings from states in $AB:C$ to states in $BC:A$ for each state in $X$ using Table \ref{table:Unitary}: 6 maps to 11 and 7 maps to $\{1,5\}$. Consequently, $\text{LEFT}(X)$ results in $BC:A=\{1,5,11\}$. The inverse operator applied to $Y$ uses the mappings from $AB:C$ to $BA:C$ given in Table \ref{table:Unitary}, so $\text{INVERSE}(Y)$ results in $CA:D = \{5\}$. The composition operator results in the union of the composition of each pairwise combination of states in its arguments. Evaluation of the compose look-up table gives the identities 
\begin{align*}
\text{COMPOSE}(1,5)&=\{1,5,11,12,17,19\}\\
\text{COMPOSE}(5,5)&=\{12,17,18,19,20\}\\
\text{COMPOSE}(11,5)&=\{17,18,19,20\}.
\end{align*}
 Therefore \begin{align*}\text{COMPOSE}(\text{LEFT}(X),\text{INVERSE}(Y))&= \\
 \{1,5,11,12,17,18,19,20\} &= Z\end{align*}

\section{Measuring Qualitative States}
\label{sec:Measurements}
Past work on qualitative mapping, particularly that using the FDC or similar representations, has characteristically taken a cognitive science approach to the problem in which the focus has been on proving the representation to be sufficient for human navigation, rather than for autonomous robotics \cite{Cohn96}. In particular, past work on the FDC has relied on the human building the map to be able to determine exact qualitative states involving all visible landmarks, but has not discussed how this might be achieved by a robot mapping an unknown area. 

This section presents a novel method of determining the possible qualitative states for landmarks visible in a camera image, without requiring knowledge of any past history or the location of the imaging point. The measurement function relies on three assumptions involving information provided by the imaging system.
\begin{enumerate}
\item Given the image, the angles to the centroids of all visible landmarks can be determined. This is equivalent to having either point-like landmarks, or landmarks with known geometries. The requirement on angle is only in the local camera frame, and there is no need for the robot to know its global orientation.
\item There is a low-level algorithm that determines the relative range ordering of visible landmarks relative to the robot. Possible methods for accomplishing this in practice include exploiting known sizes of objects, motion parallax, relative changes in object size during approach, and the fact that vertical position in an image is proportional to distance in a flat environment.
\item Landmarks are sufficiently distinctive as to be unambiguously identifiable from any orientation. Section \ref{sec:Mapping} discusses some aspects of the data association problem and how the map structure can limit the number of associations that must be considered.
\end{enumerate}

\begin{figure}[b!]
\centering
\includegraphics[width=3.5in]{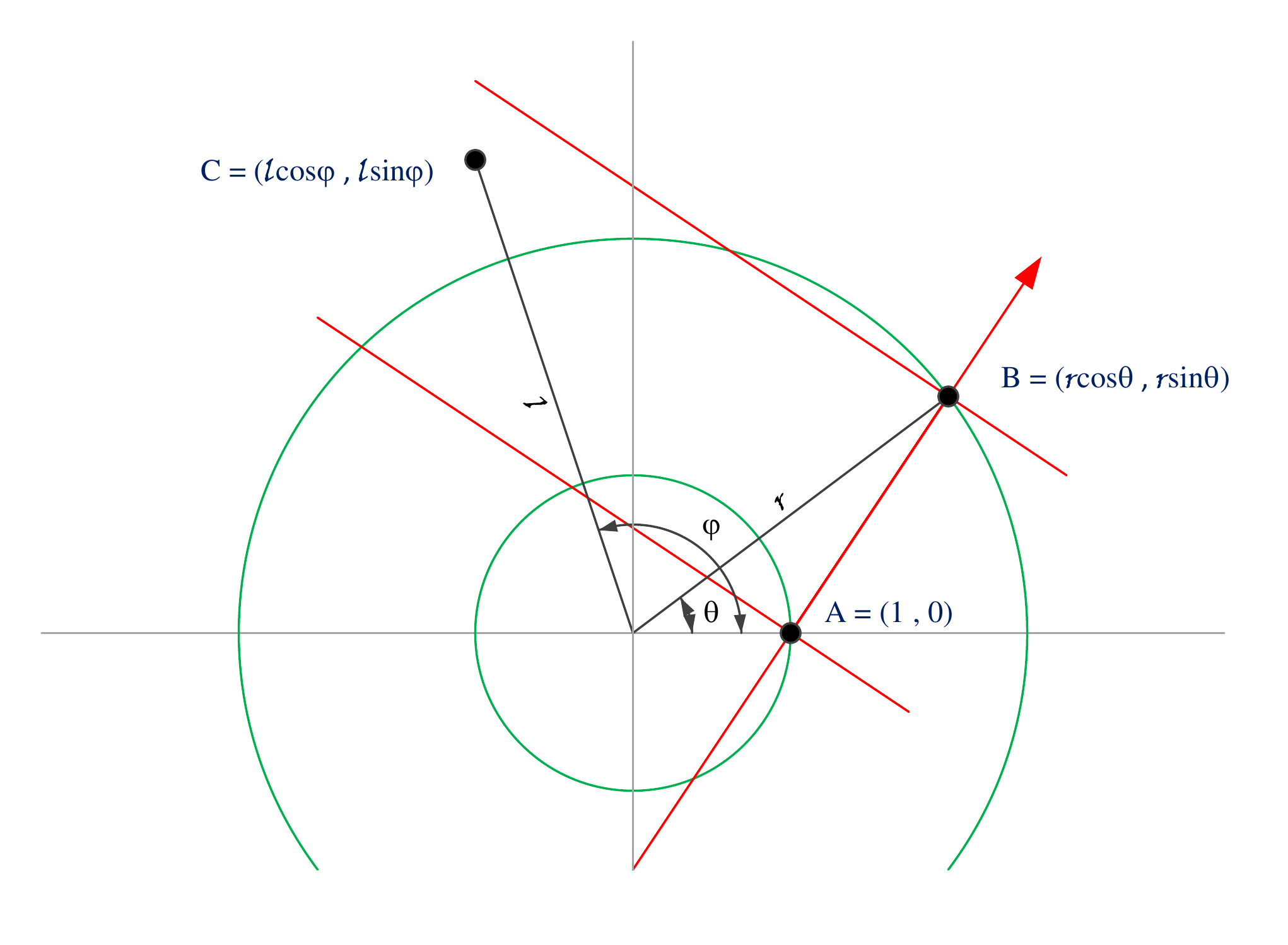}
\caption{Geometrical formulation for the problem of determining the qualitative state $AB:C$ given measurements of $\theta$, $\phi$, and the relative order of $l$, $r$, and $1$. $A$ can be freely defined to lie at $(1,0)$ as camera measurements provide only relative angle and a scaleless ordering of distances.}
\label{fig:MeasurementTrig}
\end{figure}

\begin{table*}[!tb]
\centering
\caption{EDC boundary expressions for $A=(0,1)$, $B=(r\cdot \cos(\theta), r \cdot \sin(\theta))$, $C=(l\cdot \cos(\phi), l \cdot \sin(\phi))$}
\label{table:EDCMeasurements}
\begin{tabular}{c|c}
Expression & Interpretation of Expression $<0$\\
\hline\hline
 & \\
$(\sin(\phi)\cos(\theta) - \cos(\phi)\sin(\theta))lr - \sin(\phi)l + \sin(\theta)r$ & $C$ is to the right of $\overline{AB}$\\
$-(\sin(\phi)\sin(\theta) + \cos(\phi)\cos(\theta))lr + \cos(\phi)l + \cos(\theta)r - 1$ & $C$ is in front of $A$ wrt $\overline{AB}$\\
$r^2 - (\sin(\phi)\sin(\theta) + \cos(\phi)\cos(\theta))lr + \cos(\phi)l - \cos(\theta)r$ & $C$ is in front of $B$ wrt $\overline{AB}$\\
$r^2 - 2(\sin(\phi)\sin(\theta) + \cos(\phi)\cos(\theta))lr + 2\cos(\phi)l -1 $ & $|BC| < |AC|$\\
$l^2 - r^2 -2\cos(\phi)l+2\cos(\theta)r$ & $|AC| < |AB|$\\
$l^2-2(\sin(\phi)\sin(\theta)+\cos(\phi)\cos(\theta))lr + 2\cos(\theta)r -1$ & $|BC| < |AB|$
\end{tabular}
\end{table*}

Given the bearings and range ordering for each set of three points $A$, $B$, and $C$ visible in a camera image, a measurement can be generated which consists of a list of all of the possible EDC states for $AB:C$ consistent with the observation. In a camera-centered local reference frame, the point $A$ can be defined to lie at $A=(1,0)$. Points $B$ and $C$ can then be specified as lying at $B=(r\cdot \cos(\theta),r\cdot \sin(\theta))$ and $C=(l\cdot \cos(\phi),l\cdot \sin(\phi))$, as shown in Figure \ref{fig:MeasurementTrig}, where $\theta$ and $\phi$ are the bearings measured relative to the direction of $A$, and $l$ and $r$ are the unknown ranges. The boundary conditions of EDC states may then be expressed as a series of equalities, as listed in Figure \ref{fig:EDCRegions}. These equalities are composed of the linear and quadratic expressions listed in Table \ref{table:EDCMeasurements}. Each EDC state corresponds to a set of three or four expressions being less than or greater than zero. For example, EDC state $7$ corresponds to the inequalities
\begin{align*}
(\sin(\phi)\cos(\theta) - \cos(\phi)\sin(\theta))lr - \sin(\phi)l + \sin(\theta)r>0\\
r^2 - 2(\sin(\phi)\sin(\theta) + \cos(\phi)\cos(\theta))lr + 2\cos(\phi)l -1 < 0\\
l^2 - r^2 -2\cos(\phi)l+2\cos(\theta)r < 0
\end{align*}

Determining which EDC states are consistent with camera observations can then be achieved by solving the feasibility problem of finding a point $(l,r)>0$ that satisfies both the EDC state inequalities as well as the observed ordering of $l$, $r$, and $1$. These ordering constraints are: $1-l < 0$ if $A$ is closer than $C$, $1-r < 0$ if $A$ is closer than $B$, and $r-l$ if $B$ is closer than $C$. This feasibility evaluation must be performed for each EDC state $(1-20)$. Solving these problems requires the Branch-and-Bound strategy detailed in Section \ref{sec:Feasibility}, as well as a problem-specific upper bound on $l$ and $r$ which more practically bounds the search space. Lists of EDC states consistent with the camera observation for each observed landmark triplet are passed as measurements to the graph update algorithm discussed in Section \ref{sec:Mapping}.

\section{Feasibility Detection}
\label{sec:Feasibility}
Both the calculation of measurements detailed in Section \ref{sec:Measurements} and the generation of the lookup table for Composition operator described in Section \ref{sec:Operators} require the solution of feasibility problems in either two or four variables. These problems can be formalized as determining whether there is an $x$ that satisfies a set of quadratic inequalities
\begin{equation}
\label{eq:quadratic}
x^T A_j x + b_j^T x + c_j < 0 \qquad j=1,\cdots,M\\
\end{equation}
subject to the bound constraints
\begin{equation*}
 l_b \leq x \leq  u_b
\end{equation*}
where $A_j$ are $N$-by-$N$ symmetric matrices, $b_j$, $l_b$, $u_b$ are $N$-by-$1$ matrices, and $c_j$ are scalars. As $A_j$ may be indefinite for some $j$, this problem is equivalent to a non-convex global minimization and can be shown to be NP-Hard in general. Fortunately, the small number of variables and the exploitation of the underlying geometry allows the problem to be rapidly solved using a Branch-and-Bound strategy based on that used by \citet{Maranas97}. This approach, summarized in Algorithm \ref{alg:Feasibility}, proceeds by iteratively splitting the search space into sub-rectangles, then finding a lower bound for each constraint inequality over those rectangles. If any lower bound is non-negative, then the rectangle cannot contain a feasible sub-region and is removed from the search. If all lower bounds over a rectangle are negative, then the rectangle is split in half along its longest edge and the new sub-rectangles are evaluated. 

Given the observed landmark angles $\theta$ and $\phi$, the constraint expressions listed in Tables \ref{table:EDCMeasurements} and \ref{table:Composition} are, in order of increasing complexity: linear, bi-linear, quadratic with no cross terms, quadratic with only one cross-term, or general quadratic. The exact minimum value in a rectangle can be easily found for the first four forms, while a tight lower bound can be found for the fifth. Methods for doing this are as follows:

\begin{enumerate}
\item The exact minimum value of linear and bi-linear constraints can be found by simply finding the smallest value of the constraint evaluated at each of the rectangle corners \cite{DeAngelis97}.
\item The exact minimum value of constraints with no cross-terms can be found by independently optimizing over each variable. The minimum for $x_i$ occurs at either the upper bound on $x_i$, the lower bound on $x_i$, or at $x_i=\frac{-{b_i}_j}{2{A_{ii}}_j}$ if this point lies within the rectangle.
\item The exact minimum value of constraints with only one non-zero diagonal element in $A$ can be found by one-dimensional optimizations over the corresponding variable with all other variables set to each permutation of their extreme values\cite{Vandenbussche05}.
\item Tight lower bounds can be found for general quadratic constraints by finding the minima of relaxed linear approximations as discussed by \citet{Sherali95} and summarized as follows. Dropping the subscript j, let $z = x-l_b$, $\widetilde{b}=(2 l_b^T A + b)$, and $\widetilde{c}=(l_b^T A l_b + b^T l_b + c)$, then the minimization of the right hand side of equation \ref{eq:quadratic} becomes
\begin{equation}
\label{eq:shifted}
\text{min } z^T A z + \widetilde{b}^T z + \widetilde{c} \quad s.t. \quad 0 \leq z \leq u_b-l_b
\end{equation}
The problem can be augmented by adding the nonlinear implied constraints
\begin{equation}
(g_i - G_i z) (g_j - G_j z) \geq 0 \quad \forall 1 \leq i \leq j \leq 2n
\end{equation}
where $g_i$ and $G_i$ are found by re-writing the original bound constraints 
\begin{equation}
\left[
\begin{array}{c}
	u_b-l_b - z \geq 0\\
	z \geq 0
\end{array}
\right] \equiv 
\left[
\begin{array}{c}
 g_i - G_i z \geq 0\\
i = 1, \cdots, 2n
\end{array}
\right]
\end{equation}
The augmented problem is linearized by the substitution
\begin{equation}
w_{kl} \equiv z_k z_l \quad \forall 1 \leq k \leq l \leq n
\end{equation}
The resulting linear problem is 
\begin{eqnarray}
\text{min} \; \widetilde{b}^T z + \sum_{k=1}^{n}{A_{kk} w_{kk}} + 2\sum_{k=1}^{n-1}{\sum_{l=k+1}^{n}{A_{kl}w_{kl}}} + \widetilde{c}\\
\text{subject to} \; (g_i - G_i z) (g_j - G_j z) \geq 0 \notag\\
\forall 1 \leq i \leq j \leq 2n \notag
\end{eqnarray} which can be easily solved using any off-the-shelf linear optimization routine.
\end{enumerate}

This branch-and-bound approach is guaranteed to either find a feasible solution to the constrained inequalities in Equation \ref{eq:quadratic}, or to prove that any such solution must lie within the remaining rectangles of volume less than $\epsilon$, where the value of $\epsilon$ is dependent upon the maximum iteration count. The latter case generally indicates that either there is no solution, or the the solution lies on a manifold of lower-dimensionality than the search space and thus a randomly selected point within a rectangle would be unlikely to ever exactly satisfy the constraint equations. For the feasibility problems considered in this paper, if there is a solution, it must occupy a finite volume of the search space, and an examination of the geometries involved suggests that the necessary value of $\epsilon$ should be within a few orders of magnitude of $1$. In practice, a maximum depth of 30 with an initial search rectangle of length $1,000$ gives error free results for the measurement problem in section \ref{sec:Measurements} on trials of $100,000,000$ randomly selected point combinations. Generation of the composition tables is an offline function, so a depth of $60$ was chosen to minimize the possibility of errors.

\LinesNumbered
\SetAlgoVlined
\begin{algorithm}[!htb]
add rectangle $r_0=[l_b,u_b]$ to search queue $S$\;
\label{alg:Feasibility}
\While {$S\neq 0$}{
	pop rectangle $r$ from $S$\;
	\If{ $VOLUME(r) < \epsilon$}{
		return FALSE\;
	}
	\Else{
		choose random $x^* \in r$\;
		evaluate constraints $q(x)_j=x^T A_j x + c_j^T x + d_j$\;
		\If{$q(x^*)_j < 0$,  $\forall j \in \{1,M\}$}{
			return TRUE\;
		}
		\Else{
			\For{$j \leftarrow 1$ \KwTo $M$}{
				find $\underbar{q}_j$ which lowerbounds $q(x)_j$ on $r$\;
			}
			\If{$\underbar{q}_j < 0$, $\forall j \in \{1,M\}$}{
				split $r$ into $r_l$ and $r_u$\;
				add $r_l$ and $r_u$ to $S$\;
			}
			\Else{
				continue\;
			}
		}
	}
}
return FALSE\;
\caption{Feasibility Detection}
\end{algorithm}

\section{Building Qualitative Maps}
\label{sec:Mapping}
\subsection{Map Structure}
The qualitative map generated by the algorithm presented in this section takes the form of a 3-uniform hypergraph. Each node of the graph corresponds to an observed landmark, and each edge in the graph connects three nodes and contains estimates of the qualitative states for the geometrical relationships which define their arrangement. There are six possible relationships for any three landmarks $A$, $B$, and $C$: $\{AB:C, BA:C, BC:A, CB:A, CA:B, AC:B\}$. As the inversion operator is a one-to-one mapping, the relationships $\{BA:C, CB:A, AC:B\}$ contain redundant information given $\{AB:C, BC:A, CA:B\}$, and need not be explicitly tracked in the map. Thus, an edge between the three corresponding nodes only stores estimates of $\{AB:C, BC:A, CA:B\}$. The cyclical operators which relate states in these relationships to each other are non-unique mappings, so reduction of the edge to a single relationship is not possible. Formally, the map is defined as a tuple $M=(P, E)$, where $P=\{p_1, p_2, \cdots, p_n\}$ are the nodes, and $E=\{e_{ijk}\},\: i=\{3\cdots n\}, j=\{2\cdots i\}, k=\{1\cdots j\}$ are the edges $e_{ijk}=\{p_i p_j:p_k, p_j p_k:p_i, p_k p_i:p_j\}$.

\subsection{Graph Updates}
For the following discussion, let $\overline{AB:C}$ indicate a set of states for relationship $AB:C$ stored in the graph, $\widehat{AB:C}$ indicate a measurement of the qualitative states for $AB:C$, and $\widetilde{AB:C}$ indicate temporary estimates of $AB:C$ used for intermediate steps. Information provided by measurements is propagated though the graph structure by making use of the operators discussed in Section \ref{sec:Geometry}. This procedure, equivalent to the path-consistency algorithm by van Beek \cite{vanBeek92} discussed in detail by \citet{Renz07}, operates as follows.
\begin{enumerate}
\item Given a new measurement of relationship $AB:C$, labeled as $\widehat{AB:C}$, check that nodes for landmarks $A$, $B$, and $C$ are already in the map; if not add new nodes and create new edges to all existing nodes.
\item Find the graph edge $e_{ABC}$ linking nodes $A$, $B$, and $C$.
\item Invert the states in $e_{ABC}$ if the nodes are stored in the wrong order (e.g. the measurement was $\widehat{AB:C}$ but the graph edge stored $\overline{BA:C}$).
\item Update the stored set of qualitative states $\overline{AB:C}$ by finding the set intersection with the measured states: $\widetilde{AB:C}$ = $\widehat{AB:C} \; \cap \; \overline{AB:C}$. The resulting qualitative state contains only those regions consistent with the constraints embedded in both the original value stored in the map, $\overline{AB:C}$, and the the measurement, $\widehat{AB:C}$.
\item If the intersection resulted in the set of states already stored in the map, i.e. $\widetilde{AB:C}=\overline{AB:C}$, terminate the update as the measurement contains no new information.
\item Otherwise, store the reduced set of states in the map by setting $\overline{AB:C}=\widetilde{AB:C}$.
\item Use the cyclical operators to generate pseudo-measurements $\widehat{BC:A}$ and $\widehat{CA:B}$ and update the corresponding edge states as in step 3.
\item For each qualitative state which has changed as a result of the measurement, generate new qualitative state estimates using the composition operator on all connected edges. For example, if $\overline{AB:C}$ has been updated, find all nodes $X$ which have an edge with the stored state $\overline{BC:X}$ and generate $\widehat{AB:X}=\text{COMPOSE}(\overline{AB:C}, \overline{BC:X})$
\item Treat the generated estimates $\widehat{AB:X}$ as pseudo-measurements and repeat steps 1-6 for each $X$.
\end{enumerate}

For any number and configuration of landmarks, it is guaranteed that there exists a finite image sequence which generates a fully constrained graph. Given the 2D positions, of the landmarks the imaging position of such a sequence can be predicted from evaluations of the measurement function.

\subsection{Data Association}

The mapping process described in this section relies critically on accurate measurement associations, as an incorrect association can lead to inconsistent state estimates which propagate through the graph. While the issue of consistent data association is highly problem dependent and a full discussion is beyond the scope of this work, there is one aspect of the mapping process described above that can be used to limit the number of associations to be considered. When presented with an uncertain assignment, feasibility tests can be performed on all possible qualitative states with regard to visible landmarks with good associations, just as though the landmark in question was previously unobserved. The resulting set of qualitative states can then be compared to those for existing map landmarks. Only landmarks with at least one overlapping state for each relationship need be considered for associations, as the remainder are inconsistent with the new measurement. If no possible associations remain after this step, the landmark can be safely added to the graph as a new node. If association remains unclear, the fusion of the measurement can be delayed until the map has converged further, which leads to fewer possible associations. The order in which measurements are incorporated into the map has no effect on the final map performance; the delayed fusion results in the same final map.

\section{Qualitative Navigation}
\label{sec:Navigation}

Landmark based robotic navigation can be intuitively decomposed into two distinct sub-problems: long-distance navigation between landmarks widely separated in the map, and short distance navigation between landmarks and nearby points of interest. This section focuses upon the first of these sub-problems, as there are a number of vision-based solutions to the second extant in the literature, such as visual homing \cite{Gaussier00}, place recognition \cite{Torralba03}, etc. Given the ability to reliably travel between a landmark and nearby points, long distance navigation can be achieved provided a strategy can be found to travel between the regions around any two arbitrary landmarks.

\LinesNumbered
\SetAlgoVlined
\begin{algorithm}[!htb]
\label{alg:RNG}
Given qualitative map $M=(P,E)$\;
$N = \text{sizeof}(P)$\;
Initialize $D=\{\}$, $W=\{\}$\;
\For{$i = 1 \cdots N$, $j=1 \cdots N$, $i \neq j$}
{
	add $d_{ij}$ to $D$\;
	add $w_{ij} = 0$\ to $W$\;
	\For{$k = 1 \cdots N$, $k \neq i,j$}
	{	
		$conflicts = 0$\;
		$openstates = 0$\;
		\For{\bf{all} \text{states} $s \in e_{ijk}$ }
		{
			$openstates = openstates + 1$\;
			\If{$s \in \{7,8,13,14\}$}
			{
			$conflicts = conflicts + 1$\;
			}			
		}
		\If{$openStates = conflicts$}
		{
			remove $d_{ij}$ from $D$
			remove $w_{ij}$ from $W$\;
			BREAK\;
		}
		\Else
		{
		$w_{ij} = w_{ij} + conflicts / openstates$\;
		}		
	}	
	$w_{ij} = w_{ij} / N$
}
\caption{RNG Estimation}
\end{algorithm}

\subsection{The Relative Neighborhood Graph}
\label{sec:RNG}
\begin{figure}
\centering
\includegraphics[width=3.5in]{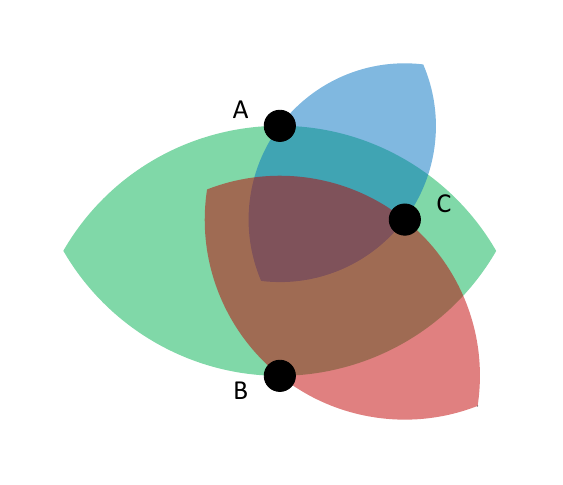}
\caption{Lunes for 3 points $A$, $B$, and $C$ which govern their neighbors in the RNG. $A$ and $C$ are neighbors, as are $C$ and $B$, but $A$ and $B$ are not neighbors as point $C$ lies within the green lune between them.}
\label{fig:lune}
\end{figure}

The navigation approach presented in this section makes use of a Relative Neighborhood Graph (RNG) of the landmarks. The RNG, as discussed by \cite{Jaromczyk92}, is a connected sub-graph of the well known Delaunay triangulation often used in computer vision, as it generates point clusters similar to those produced by humans. Landmarks are neighbors in the RNG if no third landmark appears in the lune between them. As Figure \ref{fig:lune} shows, points $A$ and $C$ are neighbors, as are $C$ and $B$, but $A$ and $B$ are not neighbors as point $C$ lies within the green lune between them.

Formally, the RNG is defined as the tuple $R = (P, D, W)$, where $P=\{p_1, p_2, \cdots, p_n\}$ are the landmark nodes used in section \ref{sec:Mapping}, $D=\{d_{ij}\}$ are edges connecting pairs of nodes $p_i$ and $p_j$, and $W=\{w_{ij}\}$ are edge costs for each edge in $D$. Estimates of the RNG can be easily extracted from the qualitative map described in section \ref{sec:Mapping} by making use of the fact that EDC states $AB:C = \{7,8,13,14\}$ correspond exactly to the lune between $A$ and $B$. An RNG edge $d_{ij}$ only exists between nodes $P_i$ and $P_j$ for which there is no third landmark in the map in any of these four states. 

In the case of an incompletely converged map, estimates of the RNG have to be realized. This is common in cases such as limited exploration of the area, reduced sensor range, and landmark occlusions during exploration. In an incomplete map, there will generally be edges that have some open states indicating that there is a landmark in the lune, and others that indicate there is not. In order to accommodate these cases, candidate RNG edges can be assigned a cost based on the number of potentially conflicting landmarks, each weighted by the fraction of open states within the lune and normalized by the total number of nodes. Edges with no conflicts have a cost of zero and are guaranteed to be true RNG edges, while edges with at least one landmark whose only open states are in the lune can be pruned from the graph. This process, summarized in Algorithm \ref{alg:RNG}, can be cheaply performed after each measurement update, adding potential RNG edges between new landmarks and all existing landmarks in the map, then pruning them away based on the graph updates. As RNG edge estimates depend only on determining if the lunes of landmark pairs contain other landmarks, the convergence rate is bounded above by that of the qualitative map. However, in practice, close approximations to the true RNG are often found early in the mapping process, long before the qualitative map is fully constrained.

\subsection{Graph-based Navigation}
\label{sec:GraphNavigation}

The navigation approach presented in this section assumes that local planners can reliably operate with the Voronoi cells of each landmark (the locus of points closer to the selected landmark than to any other). The goal is then to enable travel between Voronoi cells of the landmark closest to the start and goal points. This can be achieved using the limited sensors used in Section \ref{sec:Measurements} and the RNG described above, assuming that landmarks are visible from adjacent cells. A simple, but effective, navigation strategy is as follows:
\begin{enumerate}
\item Given the start and goal points START and END
\item Find $p_s$ and $p_e$, the closest landmarks to START and END respectively
\item Use a graph search algorithm to find the shortest sequence of intermediate landmarks connected by RNG edges between $q_s$ and $q_e$. If the map is well constrained (i.e. the RNG estimate is close to the true RNG), then Dijkstra's algorithm is sufficient. In less constrained cases a weighted approach, such as $A^*$ or $D^*$ is likely to be more effective, using the RNG edge costs as a heuristic to be added to a fixed separation-based distance costs. This biases the search towards paths along the RNG edges least likely to be pruned away by new measurements and towards edges that are most likely to be correct.
\item Drive towards the first landmark in the search path until the rover enters its Voronoi cell, as detected by the relative range orderings of observed landmarks.
\item Remove the current landmark from the search path and drive towards the second landmark until you reach its Voronoi cell.
\item Repeat steps 4 and 5 until the rover has entered the Voronoi cell around $p_e$
\item Use a local planner to drive to END
\end{enumerate}

\begin{figure}
\centering
\subfloat[]{\includegraphics[width=3in]{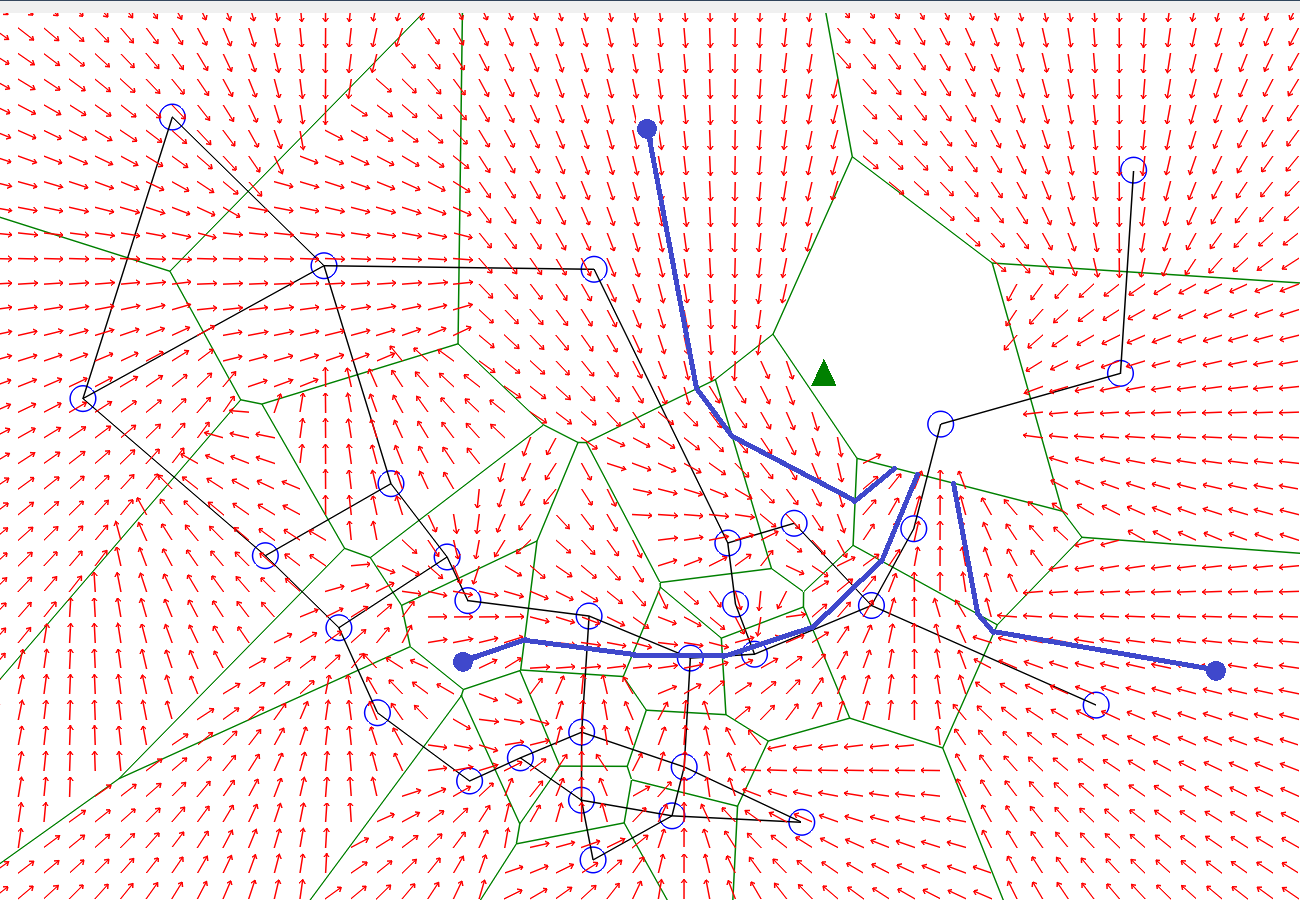}\label{fig:flow1}}\\
\subfloat[]{\includegraphics[width=3in]{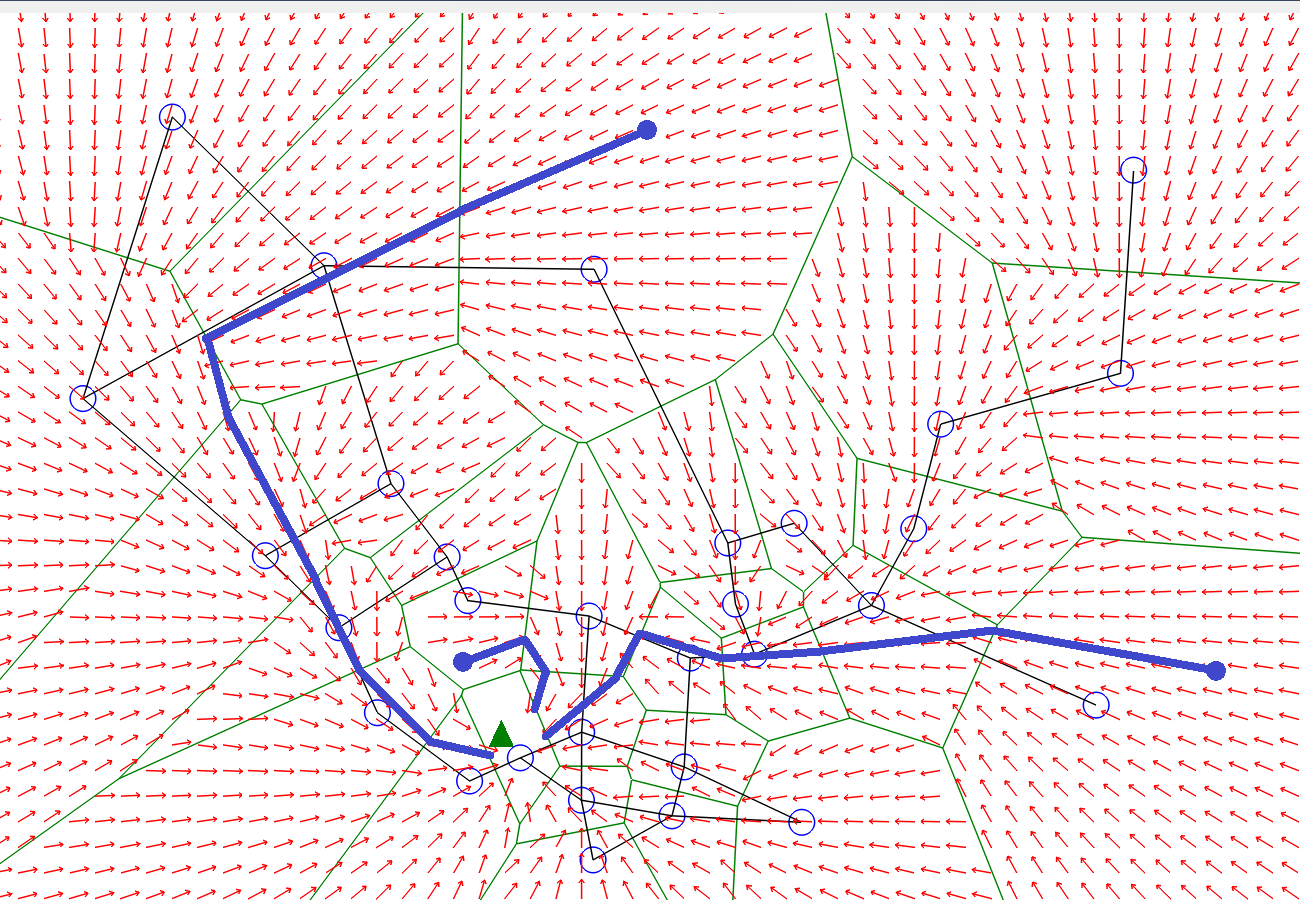}\label{fig:flow2}}
\caption{Example flow fields along which a robot would travel using the navigation strategy discussed in section \ref{sec:GraphNavigation}. Open circles indicate landmark positions, green lines indicate the borders of Voronoi cells for each landmark, black lines indicate the RNG estimate used for navigation. Red arrows indicate the direction of motion calculated at each point for a robot traveling to the Voronoi region containing the green triangle. Blue lines indicate trajectory of a robot starting from closed circles at three random starting locations.}
\label{fig:flowFields}
\end{figure}

Figure \ref{fig:flowFields} shows flow fields for two randomly selected END locations, along with examples of rover trajectories from random START locations. These trajectories show that the navigation approach generates paths that trade distance optimality for a guarantee on reaching the goal, given the assumptions on landmark visibility states in Section \ref{sec:Measurements}. As the landmark distribution approaches uniformity, distance in the RNG becomes a better proxy for metrical distance, and the difference between trajectories generated by the above strategy and optimal paths decreases. Critically, the navigation approach has no control requirements other than that the robot can always make forward progress towards visible landmarks, and it only requires sensing of local landmarks in order to achieve long-distance objectives. 

\section{Mapping Results}
\label{sec:Results}

\begin{figure*}[!tb]
\centering
\subfloat[][Removed EDC States]{\includegraphics[width=3in]{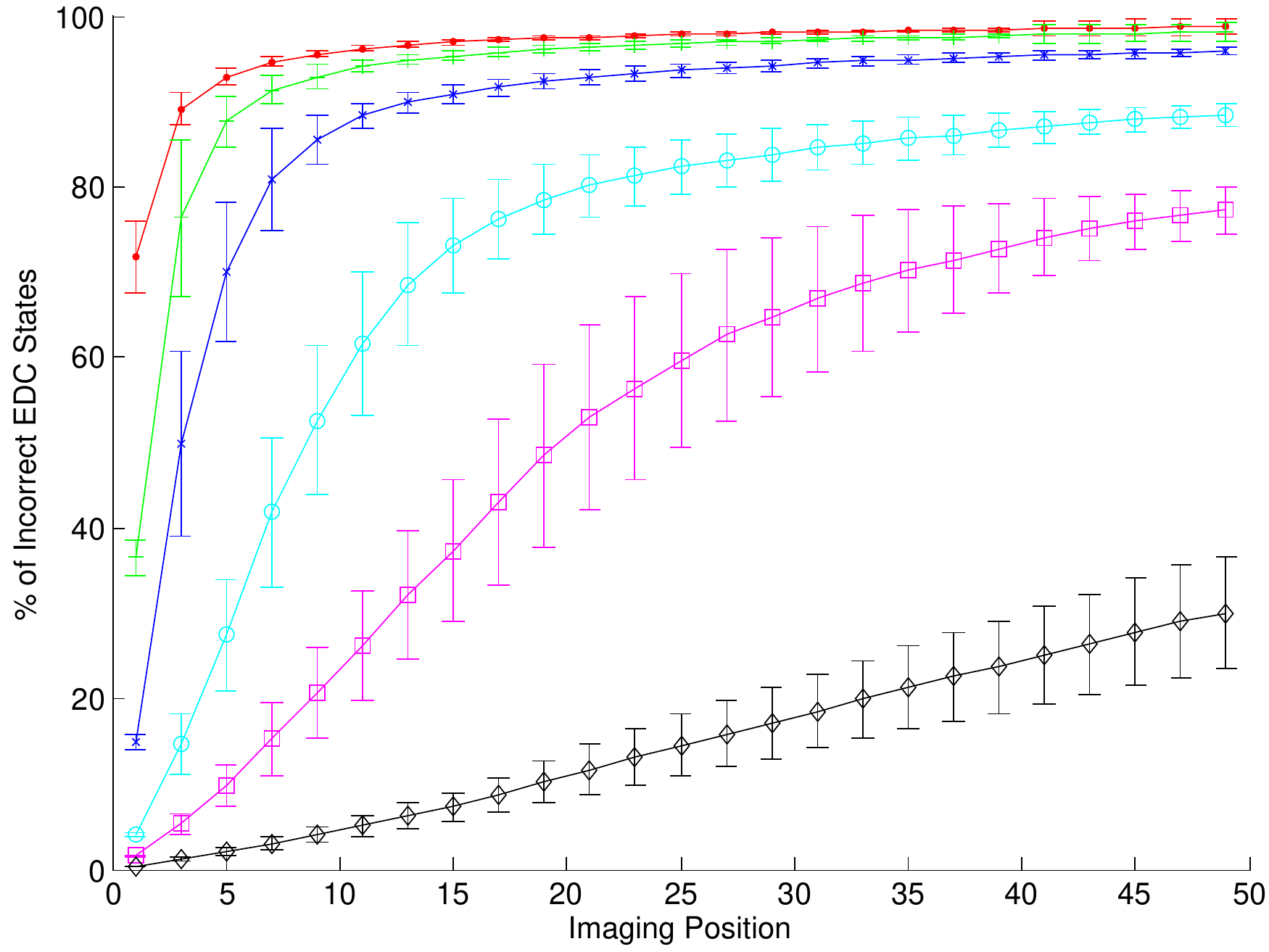}\label{fig:MCResultsA}}\qquad
\subfloat[][Non-Adjacent Open EDC States]{\includegraphics[width=3in]{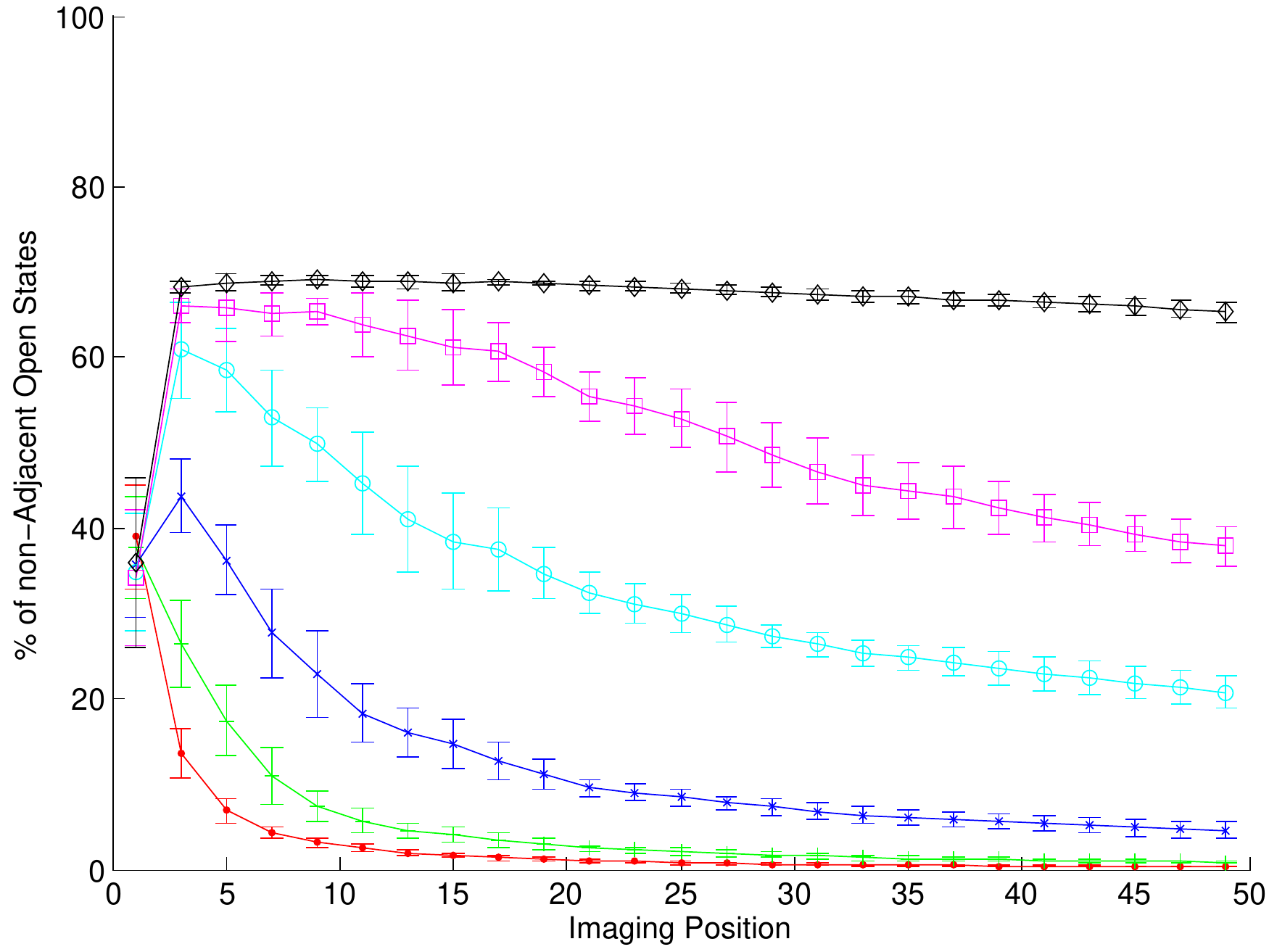}\label{fig:MCResultsB}}\qquad
\subfloat[][RNG Edge Cost]{\includegraphics[width=3in]{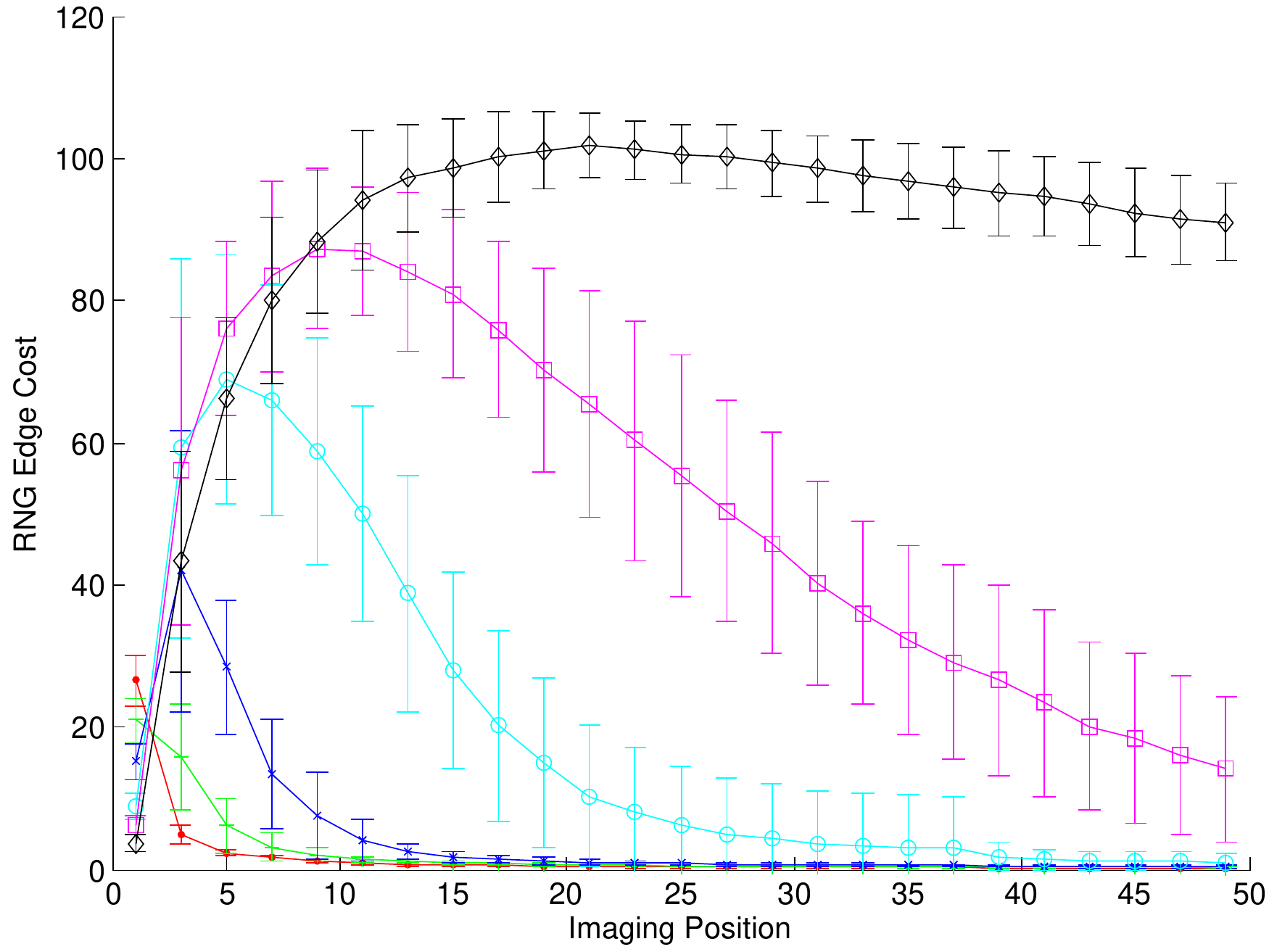}\label{fig:MCResultsC}}\qquad
\subfloat[][Computation Time]{\includegraphics[width=3in]{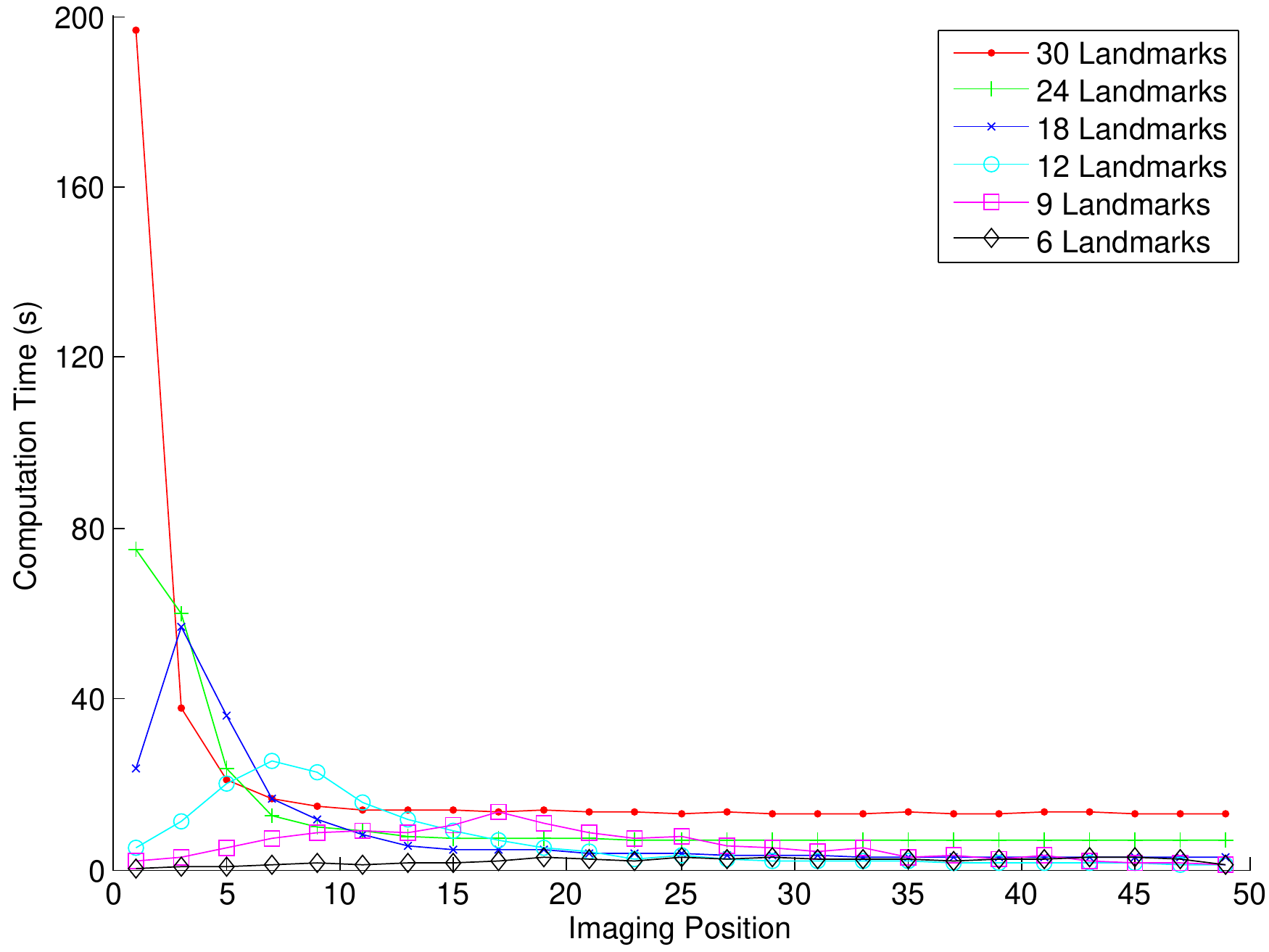}\label{fig:MCResultsE}}\qquad
\caption{Monte-Carlo performance of the QRM algorithm as measurements are incorporated into the map, as a function of the number of $n$ closest landmarks used at each imaging position. The legend in (d) applies to all plots. (a) means and standard deviations of the cumulative percentage of incorrect EDC states that have been removed from the map due to being inconsistent with the observed measurements. (b) means and standard deviations of the percentage of open EDC states which are not adjacent to the true state. (c) means and standard deviations of the total cost of all remaining RNG edges that have potentially conflicting nodes. (d) mean computation times for each measurement update. The relative deviations are not shown for the sake of clarity, but averaged between 15\% and 30\%.}
\label{fig:MCResults}
\end{figure*}

\begin{figure*}[t!]
\centering
\includegraphics[width=7in]{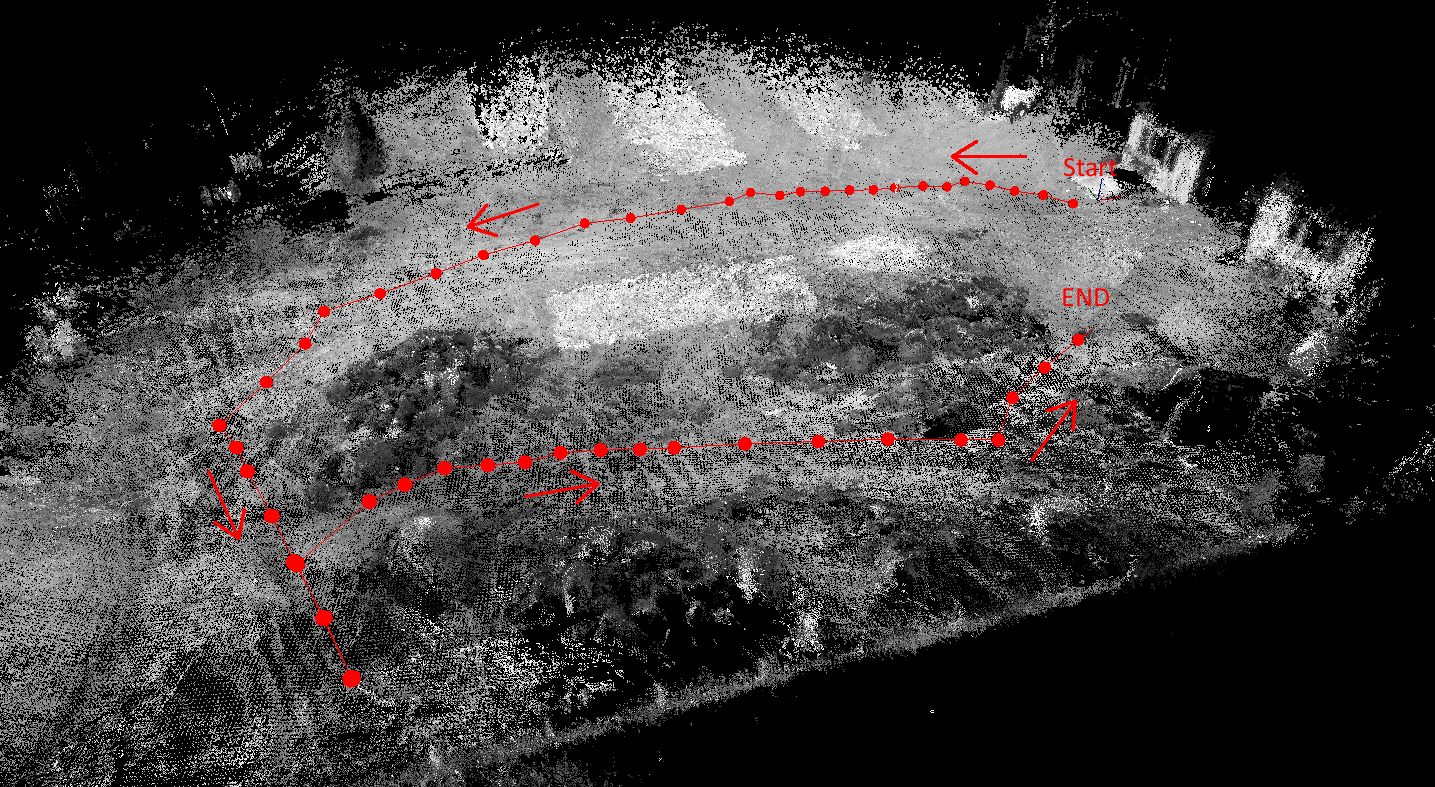}
\caption{3D reconstruction of the JPL Mars Yard. The pointcloud was generated from stereo panoramas taken at the imaging points denoted by red circles, stitched together using the NDT \cite{Magnusson07} and LUM \cite{Lu97} algorithms implemented in the PointCloud Library. Landmarks include medium sized rocks such as those in the image center as well as similarly sized objects such as the generators in the upper left and right corners.}
\label{fig:MarsYard}
\end{figure*}

This section discusses a series of Monte-Carlo simulations designed to test the QRM algorithm, as well as experimental results of a mapping the JPL Mars Yard.

\subsection{Map Evaluation Metrics}

There are three primary measures for evaluating the quality and convergence of a relational map. The first is the number of incorrect EDC states that have been removed from the graph by measurement updates, expressed as a percentage of the total number of possible states. The second performance metric is the number of map edges that have been fully constrained, i.e. have only one remaining open state for each of the relationships $AB:C$, $BC:A$, and $CA:B$, again expressed as a percentage of the number of edges in the final map. The third metric is the sum of the edge costs in the RNG, $\sum_i{\sum_j{w_{ij}}}$, which indicates the degree of confidence in the RNG estimate after each measurement.

\subsection{Monte Carlo Simulations}
\label{sec:MonteCarlo}

This section presents the results of a set of Monte Carlo tests designed to illuminate properties of the QRM algorithm defined above. The simulation operates on a set of specified landmark and imaging locations. At each imaging point, the simulation generates measurements of all detected landmarks using the process described in Section \ref{sec:Measurements}. A qualitative map is built by combining measurements from each imaging point sequentially using the approach detailed in Section \ref{sec:Mapping}. A total of 100 Monte-Carlo simulations were run to examine the general trends of the QRM algorithm for arbitrary map configurations. For each run, 30 landmarks were randomly generated from a uniform distribution in a square map. Simulated measurements were then taken from 50 randomly chosen imaging locations and combined into a qualitative map using the method described in Section \ref{sec:Mapping}. 

The QRM algorithm performs best when the robot is able to see all of the landmarks in each image, as this allows measurements extracted from each image to potentially add new constraints between all landmarks directly, without needing to rely on less precise information propagated through the graph. This situation is not generally true in practice, as landmark visibility is reduced by both range and occlusions. In addition, for computational reasons the number of landmarks used at each location may be limited. The effect of sensor limitations was tested by evaluating algorithm performance using only closest $n$ landmarks to the rover at each imaging point. For uniformly distributed landmark maps, this measurement restriction is equivalent to imposing a maximum sensor range. Results for these simulations are shown in Figure \ref{fig:MCResults} for values of $n$ ranging from $6$ to $30$. 

Figure \ref{fig:MCResultsA} shows statistics for the number of EDC states removed from the graph as a percentage of the total number of possible incorrect states for 30 landmarks. Each plotted line corresponds to a different number of landmarks used for measurement updates, as indicated in the legend in Figure \ref{fig:MCResultsE}. The overall trend is for a rapid initial pruning of the incorrect states, followed by a slow tapering as the remaining states are removed. The initial measurement of a landmark triple is always able to remove at least half of the potential states, as seen by the jump at image position $1$. While the initial measurements are able to greatly reduce the number of unconstrained states, the system requires the repeated observation of landmarks from different orientations in order to constrain any landmark triplet to a single state. Consequently, a randomly selected imaging location becomes progressively more unlikely to provide additional constraints on more than a few landmark relationships, manifesting as a slow convergence towards the fully constrained case. Convergence is fastest for the case in which all landmarks are  measured (the red 30 landmark line); however the cases using the nearest 18 and 24 landmarks perform nearly as well by the end of the simulations, despite a slower start. In contrast, convergence slows dramatically when less than half of the landmarks, i.e. $(n<15)$, are used in each imaging measurement.

Figure \ref{fig:MCResultsB} plots statistics for the percentage of open EDC states in the map which are not adjacent (sharing a boundary edge or vertex) to the true state of the associated landmark triple. If open EDC states are uniformly distributed, this value ranges between $45\%$ and $90\%$, depending on the exact geometry of the map. Values lower than $45\%$ indicate that the open states are clustered around the true states, i.e. the map ambiguities are primarily between adjacent states, while values above $75\%$ indicate that the remaining ambiguities are between distant states. The results in Figure \ref{fig:MCResultsB} show that for the $18$, $24$, and $30$ visible landmark cases, the map ambiguities are quickly reduces to states close to the true state, while the $6$ and $9$ landmark cases stay within the uniform range.

Figure \ref{fig:MCResultsC} plots statistics for the total edge cost of the RNG estimates after each update. The initial peak in total edge costs followed by a convergent tail corresponds to adding landmarks to the graph, and consequently additional edges, faster than there is enough data to remove existing incorrect edges. For the cases where more than half of the landmarks are used at each step, the RNG estimates clearly converge faster than either the map qualitative states or edges. This trend is unsurprising, as the RNG estimates depend on determining only whether landmarks lie within one of the four lune states, so a great deal of ambiguity can still be present in the map as a whole even after the RNG has converged.

Figure \ref{fig:MCResultsE} shows the mean computation time required for each measurement update; the relative deviations are not shown for the sake of clarity, but were typically $15 - 30\%$. Simulations were performed using unoptimized C\# code running on a Pentium Xeon at 2.5GHz. At each step, computation costs are dominated by the number of feasibility tests that must be performed in order to generate measurements, which depends on the number of landmarks observed and the number of open EDC states in the map. When all, or nearly all, landmarks are seen in every image, the peak computation time occurs in the initial measurements, as every landmark triplet must be checked, and every EDC state is open. A power-law analysis shows that that this peak cost scales as $O(n^3.5)$ with the number of landmarks. When a small number of landmarks are seen in each image, the initial cost is greatly reduced, as only a few triplets need to be checked, and the map itself contains fewer edges. Results suggest that if limited computation is available, the map may be initialized using only a subset of the visible landmarks, and then the number of landmarks used increased as the map becomes more constrained. Alternatively, the fusion of measurements for less important landmarks may be  delayed until additional computing resources are available. Although this may reduce the accuracy of the map initially, the final performance will be the same regardless of the order in which measurements are fused.

\subsection{Data-Driven Simulation}
\label{sec:MarsYardResults}

This section presents a scenario designed to illuminate some of the properties of the Qualitative Mapping and Navigation algorithms developed in this paper. The platform used was a 6-wheeled rocker-bogey frame with a mast-mounted stereo camera functionally equivalent to the two Mars Exploration Rovers (MER), Spirit and Opportunity. The experiment objective was to construct a qualitative map of a set of rock fields in a Mars-like environment, with Mars-like hardware and operations. The rover was driven through the field, stopping to take panoramic images every 1-2 meters of travel. Landmarks measurements were extracted from these images using the method presented in Section \ref{sec:Measurements} and combined using the mapping algorithm described in Section \ref{sec:Mapping}.

\begin{figure}
\centering
\includegraphics[width=3.5in]{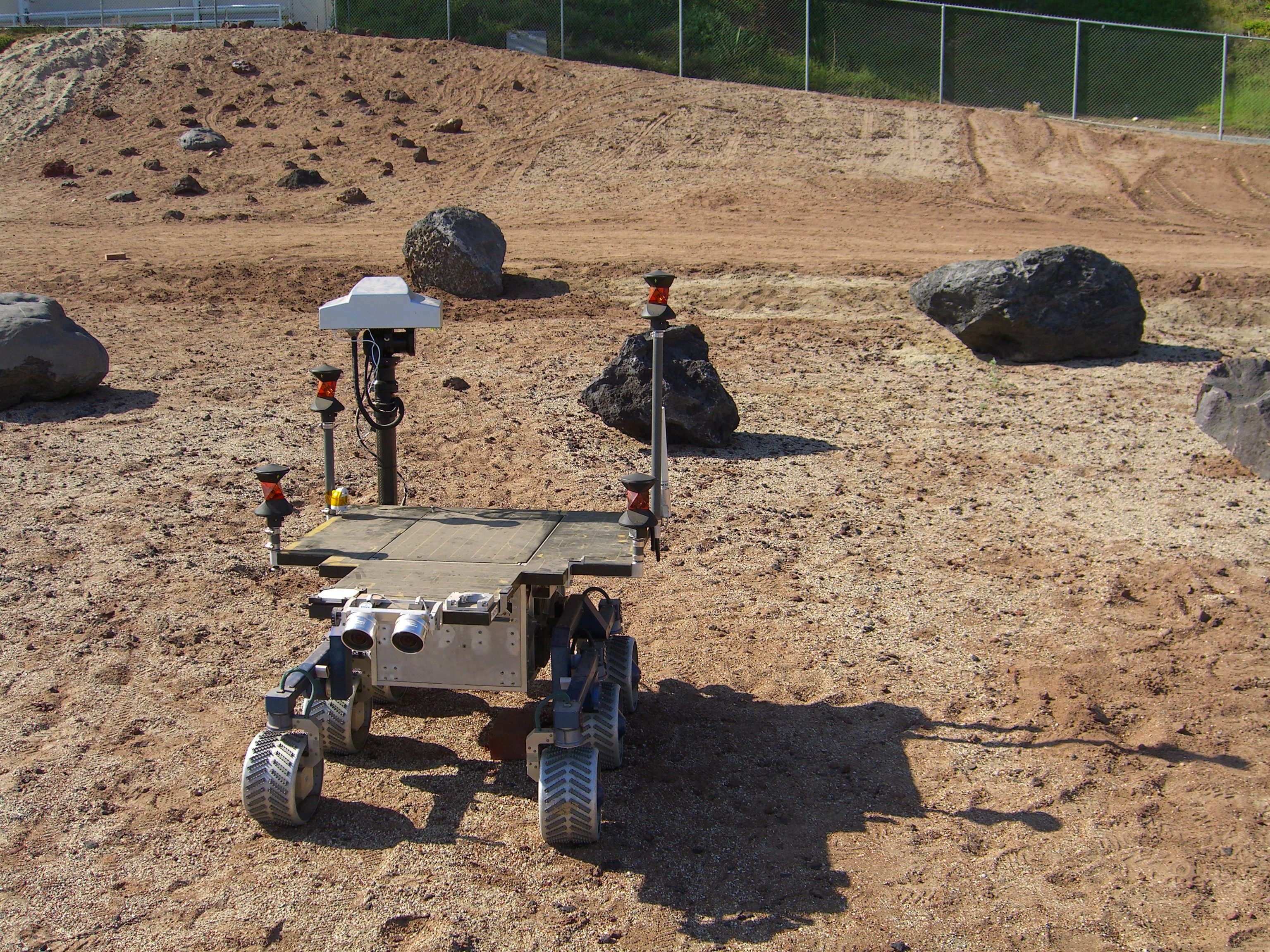}
\caption{The FIDO research rover operating in the JPL Mars yard. The 3D reconstruction of the area shown in Figure \ref{fig:MarsYard} was performed using images taken by a stereo camera pair located on the sensor mast.}
\label{fig:FIDO}
\end{figure}

As the Mars Yard data did not include the necessary parallax information for range ordering of landmarks in each image, these measurements were generated using the true position of the rover and landmarks at each imaging point. This process relied on extracting landmark and rover positions from a 3D reconstruction of the environment overlaid with the rover trajectory and imaging locations, as shown in Figure \ref{fig:MarsYard}. Stereo ranges were computed using the approach presented in \cite{Goldberg02} and converted into robot-centered pointclouds containing position and intensity data. The set of clouds from images taken at a single position were aligned using mast attitude measurements, then refined using the Normal Distribution Transform approach of \citet{Magnusson07}. These panoramic clouds were initially aligned using position estimates from rover odometry, and fused into the final map using the batch alignment method described by \citet{Lu97}. The rover traversal formed a loop through the Mars yard, which created a significant overlap in points between the first and last imaging position that was exploited to construct a circular graph of correspondences in order to minimize position drift.

\begin{figure}
\centering
\subfloat[][Removed EDC States]{\includegraphics[width=3.5in]{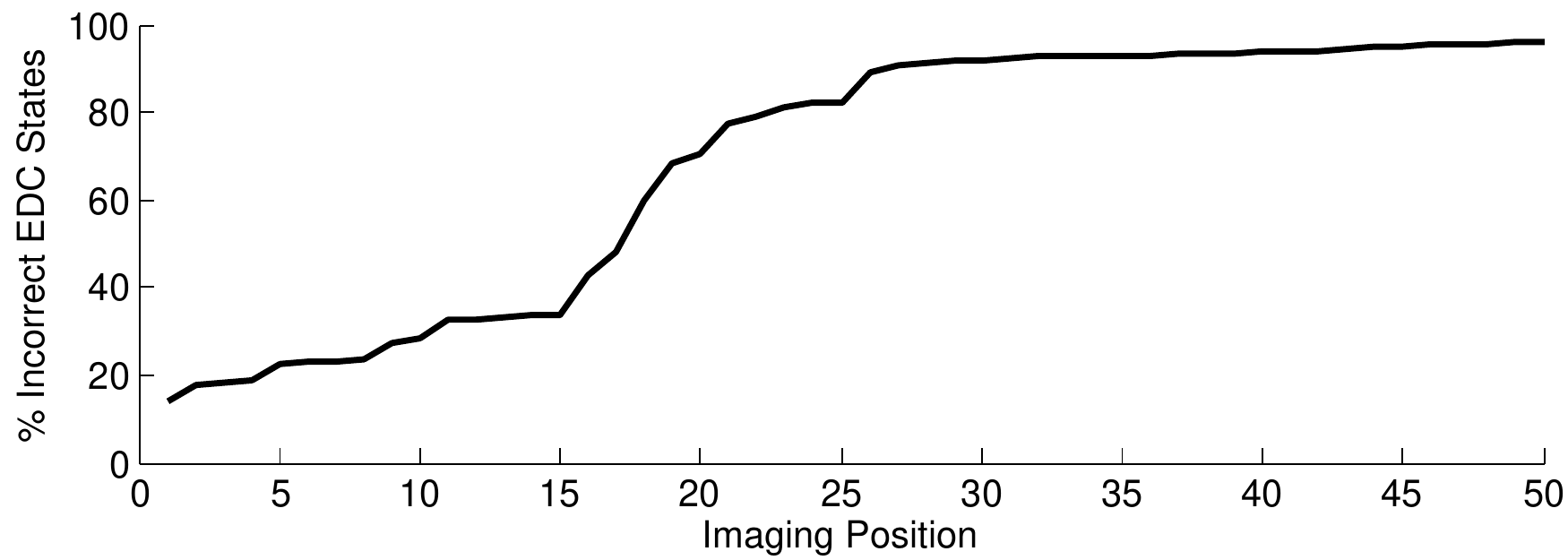}\label{fig:MYResultsA}}\\
\subfloat[][Non-Adjacent Open EDC States]{\includegraphics[width=3.5in]{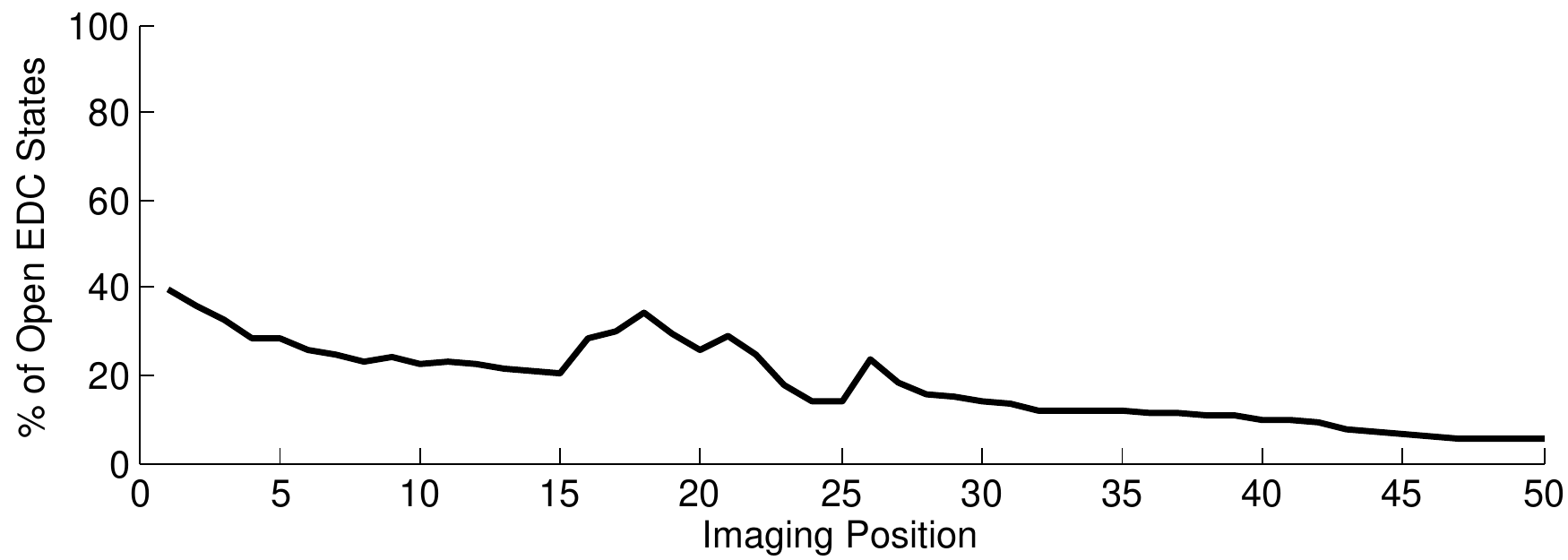}\label{fig:MYResultsB}}\\
\subfloat[][RNG Edge Cost]{\includegraphics[width=3.5in]{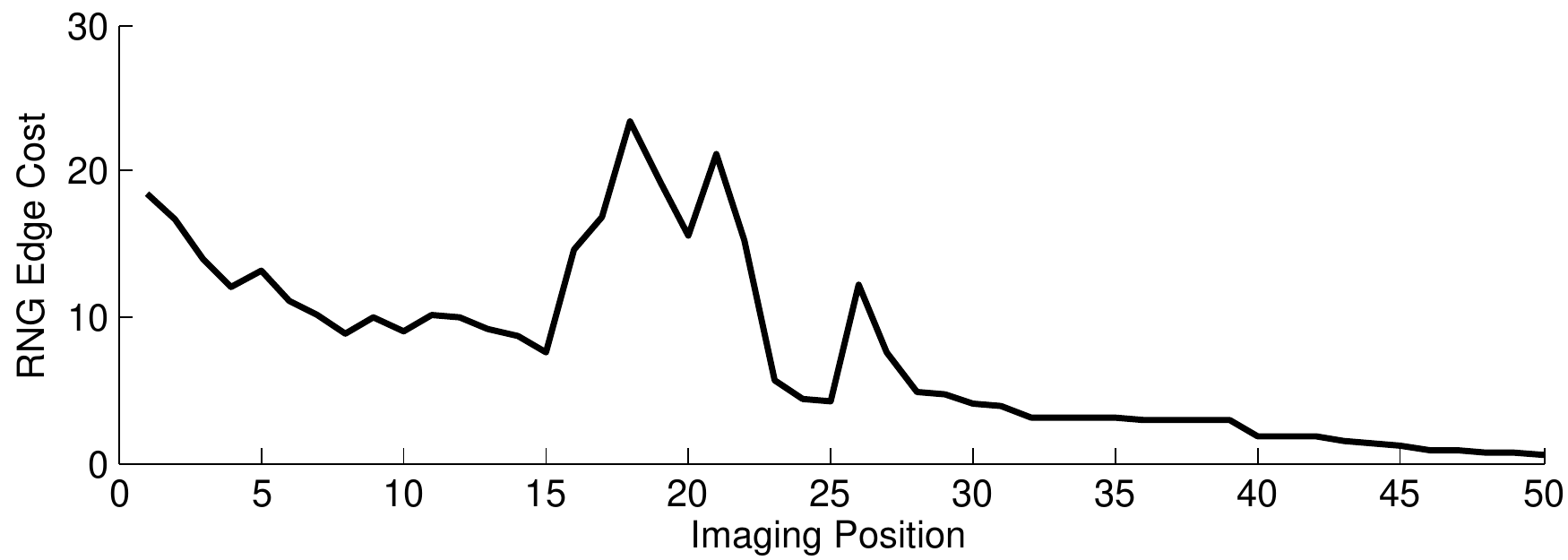}\label{fig:MYResultsC}}
\caption{Mapping performance of the QRM algorithm for the FIDO rover traversal of the JPL mars yard described in section \ref{sec:MarsYardResults} and shown in Figure \ref{fig:MarsYard}. At each step, measurements of the 18 landmarks nearest the rover were taken. (a) the cumulative number of EDC states removed from the map after each imaging point, as a percentage of the number of EDC states for a 30 landmark map. (b) the percentage of open EDC states that are not adjacent to the true state. (c) the total cost of the RNG edges extracted from the map at each step, where cost is equal to the number of conflicting states as a fraction of the total number of open states.}
\label{fig:AlgPerformance}
\end{figure}

The 30 most visually distinct objects of appropriate size in the environment were manually selected as landmarks for the mapping algorithm. These primarily consist of medium sized rocks in one of the clusters seen in the center of Figure \ref{fig:MarsYard}, but also a few man-made objects such as the generators seen in the corners of the field. At every imaging location the rover stopped and captured a panorama using the mast-mounted cameras. Landmarks were manually extracted from the left camera images and compared against the reconstruction for data association. The nearest 18 landmarks were then used to perform map updates. While the mapping algorithm described previously was run on this data set offline, a desktop computer was able to construct the map in real-time.

\begin{figure*}[!tb]
\centering
\subfloat[][After 1st Measurement]{\includegraphics[width=3in]{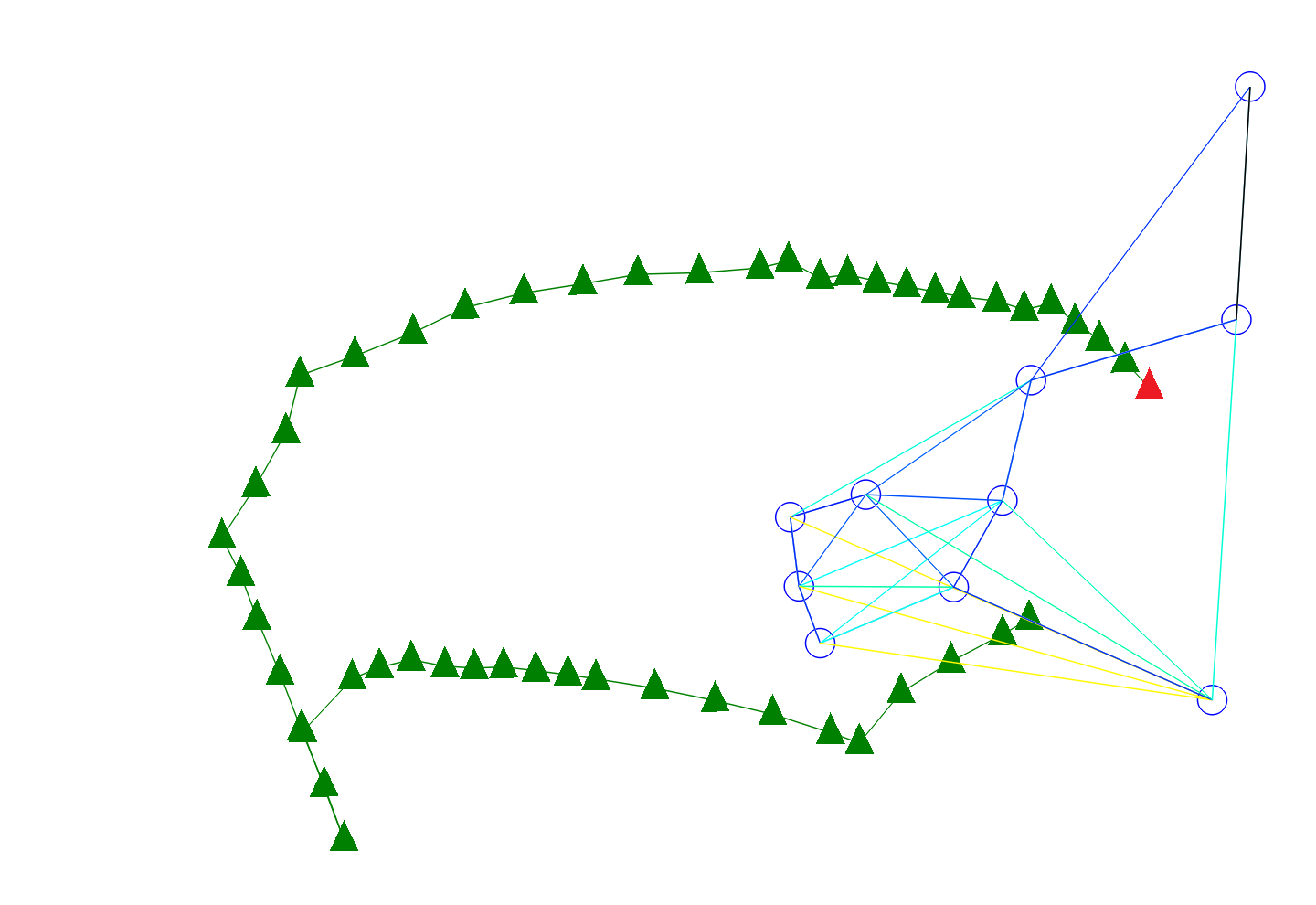}\label{fig:RNG1}}\qquad
\subfloat[][After 15th Measurement]{\includegraphics[width=3in]{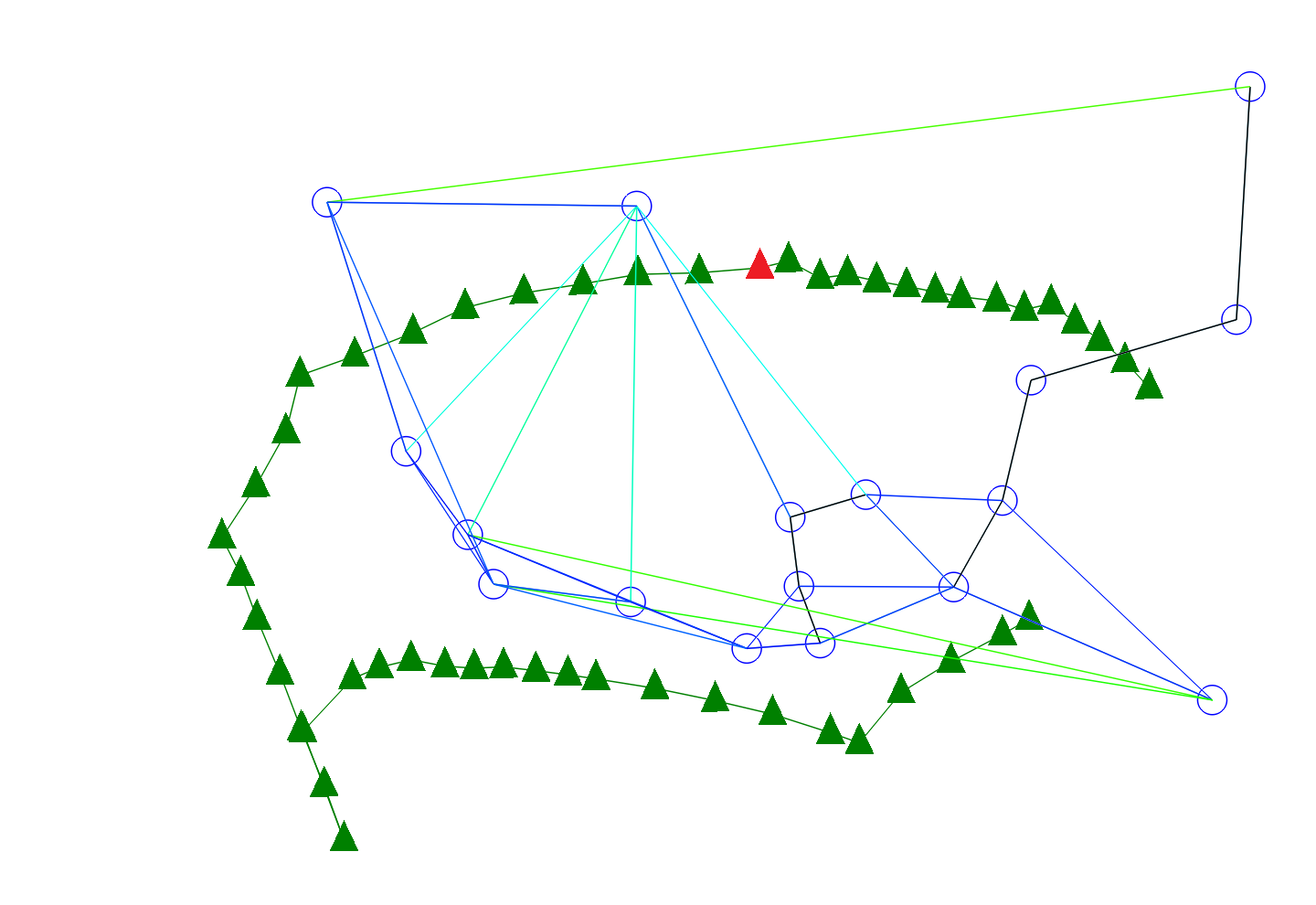}\label{fig:RNG2}}\qquad
\subfloat[][After 30th Measurement]{\includegraphics[width=3in]{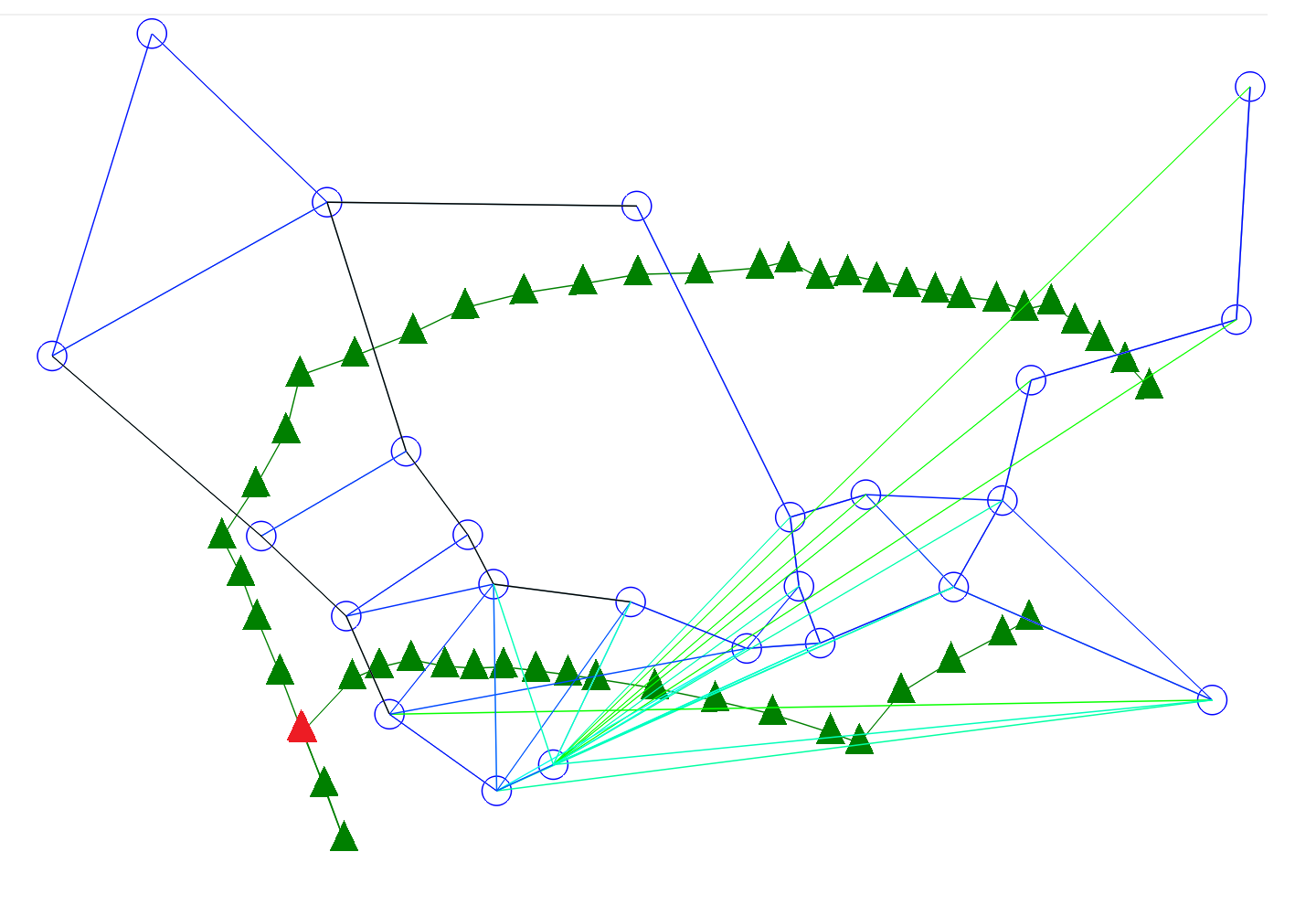}\label{fig:RNG3}}\qquad
\subfloat[][After 50th Measurement]{\includegraphics[width=3in]{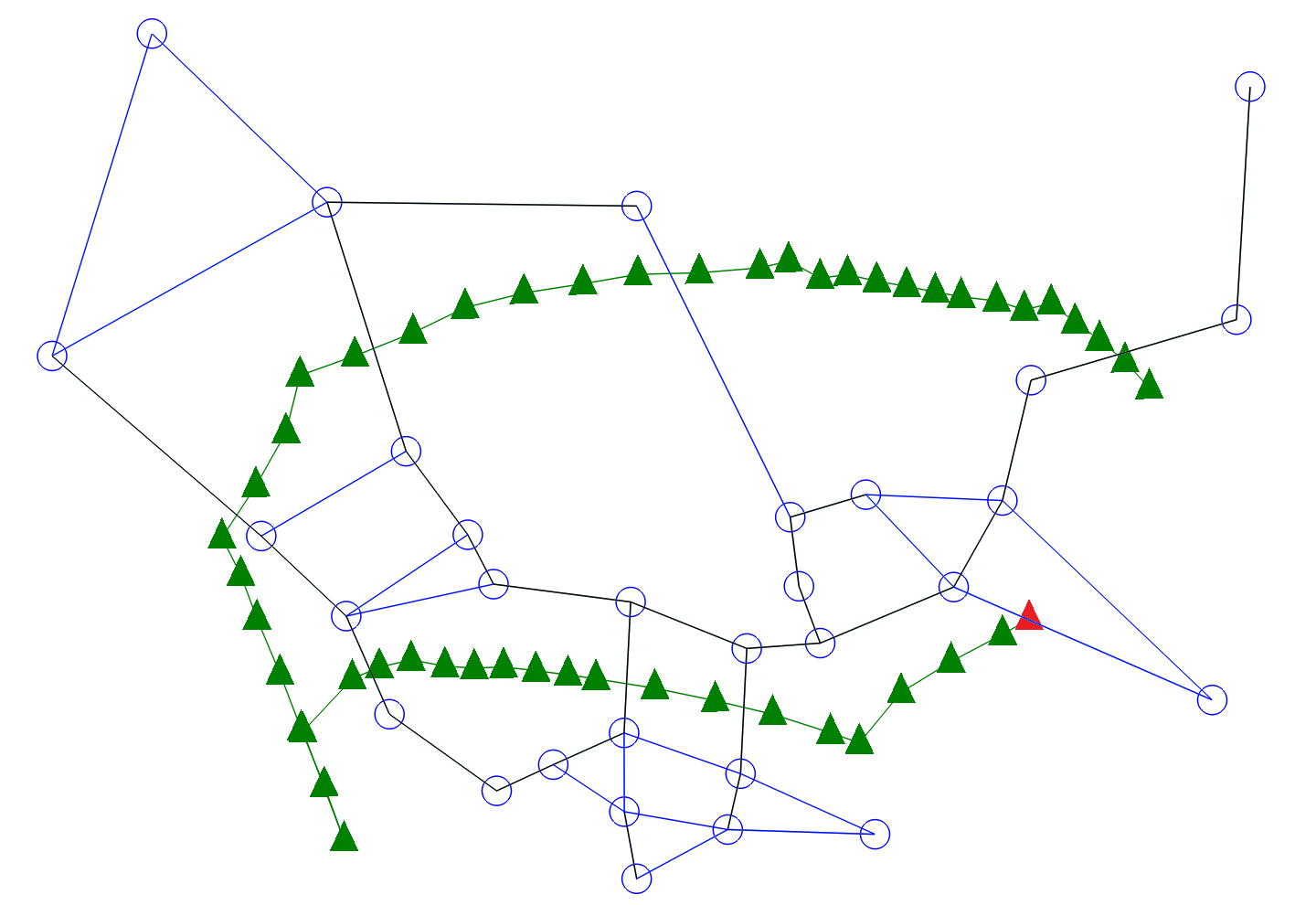}\label{fig:RNG4}}\qquad
\caption{Evolution of the Relative Neighborhood Graph (RNG) extracted from the qualitative map generated for the rover traversal described in section \ref{sec:MarsYardResults} and shown in Figure \ref{fig:MarsYard}. Blue circles indicate the 2D locations of landmarks already observed and included in the map, green triangles indicate the imaging locations, with the rover starting in the upper right corner, the red triangle indicates the most recent imaging location, and lines indicate RNG edges. RNG edges are colored according to the edge weights $w_{ij}$ discussed in Algorithm \ref{alg:RNG}, with black indicating that $w_{ij} = 0$ and then ranging from dark blue to red as $w_{ij}$ ranges from $0$ to $1$.}
\label{fig:MarsYardRNG}
\end{figure*}

Mapping results for this experiment are shown in Figure \ref{fig:AlgPerformance}; as a comparison to the prior simulation results, the Mars yard experiment results can be compared to the blue lines in Figure \ref{fig:MCResults}(a-c). The most striking features when compared against the Monte-Carlo results is the slower convergence of map states and the distinct sigmoid shape of the plot in Figure \ref{fig:MYResultsA}. This can be attributed to two distinguishing characteristics of a realistic traversal. The first is that the FIDO rover explored a smaller fraction of the total map, measured in terms of rover footprint, than a randomly selected set of imaging locations does. This means that the range of novel viewpoints is more limited than the total waypoint count alone suggests. The second, related, characteristic is that sequential images are highly correlated in this experiment, and thus provide much less additional information than would a new image taken from a random point on the map. These effects are particularly apparent in the first few images, as the random points used in the Monte-Carlo tests are likely to be far apart, and thus provide very different views of the scene, while the first few waypoints in the FIDO traverse are close together. The effects of these differences are less apparent toward the end of the simulations, as in both cases, new images are taken near previously visited locations and provide a limited amount of new information regarding landmark qualitative states.

The RNG performance shown in \ref{fig:MYResultsC} shows equivalent final performance, reduced peak error, but a higher average error in the middle of the traverse than seen in the Monte-Carlo results. This is a direct result of the limited sensor range and sequential measurements. As the traverse imaging points are close spatially, the system tends to acquire the measurements necessary to constrain the RNG faster than observing new landmarks and adding them to the graph. This results in a slowly growing, but well constrained RNG estimate at each step. This progression can be best seen through the snapshots of the RNG shown in Figure \ref{fig:MarsYardRNG}, taken after the 1st, 15th, 30th, and 50th (final) measurements are incorporated into the graph. In these plots, RNG edges are removed from consideration if any landmark is confirmed to lie within the lune (i.e. there is some landmark $C$ s.t. $AB:C$ has no components except for members of the set $\{7,8,13,14\}$. These figures illustrate that as the RNG grows, only a few long-distance (and thus, for this case, incorrect) edges are maintained, and that these edges tend to have a high edge cost. When a new landmark is observed, as occurs in \ref{fig:RNG3}, there is sufficient information stored in the map to restrict its incorrect connections to only a few nodes, and these erroneous edges are quickly pruned away by subsequent measurements.

Total computation time for generating the qualitative map and updating the RNG estimate at each step was $250$ seconds. For comparison, the FIDO rover required approximately six hours of continuous operation to perform the traverse, stopping every $1-2$ meters to collect a panoramic image. The speed of this process was primarily limited by the inefficient method used to gather panoramic images and the rovers low top speed of $9$cm/s.

\section{Conclusion} 
\label{sec:Conclusion}

A novel landmark-based mapping and long-distance navigation approach using qualitative geometry has been presented. The problem of long-distance operation of robots in sparse, unconstrained environments is considered, using the robotic exploration of Mars as an example applications. The algorithms generate and operate on graphical networks which store constraints on qualitative geometrical relationships between triplets of landmarks in the map based on limited sensor measurements. The underlying qualitative representation of these relations, termed the Extended Double Cross (EDC), defines constraints on the qualitative distances between landmark triplets as well as their qualitative angles. This mapping approach performs a form of qualitative triangulation based on angle measurements and estimates of the relative range orderings of visually distinctive landmarks; these measurements are consistent with current Mars-rover sensing technology. Both the measurements and the offline generation of lookup tables for converting between states make use of a Branch-and-Bound approach to determining the feasibility of sets of non-convex quadratic inequalities. The hypergraph constructed by this algorithm provides a description of the landmark geometries which is invariant under translation, rotation, and uniform scaling transformations. Robot navigation objectives can be expressed in terms of the intersecting regions formed by the EDC state boundaries associated with the landmarks; for example `stay to the right of points A and B', can be re-expressed in terms of desired qualitative states with respect to the map graph. An example navigation strategy was presented which uses estimates of the landmark Relative Neighborhood Graph (RNG) extracted from the qualitative map in order to find paths between the Voronoi regions of arbitrary landmarks. 

The asymptotic behavior of the mapping system was evaluated using Monte-Carlo simulations of randomly generated maps with 30 landmarks, and simulated rovers which utilize a varying number of the closest landmarks at uniformly distributed random imaging positions. Results show that while map convergence rates are closely linked to the number of landmarks simultaneously observed, the system has similar asymptotic performance when at least half the landmarks are used at each step, and with greatly reduce computational requirements compared to simultaneous observations of all 30 landmarks. The results also demonstrate that computational cost is strongly tied to the presence of new landmarks in a measurement, so that an incremental map-building strategy is preferred to a batch approach. The qualitative mapping system was also evaluated using data-driven simulations based on a traversal of the Jet Propulsion Laboratory Mars Yard by the FIDO research rover. A 3D reconstruction of the yard was used to determine the true rover trajectory and landmark locations. Results show that although overall map convergence was slower than the Monte-Carlo results with random imaging locations, due to correlations between sequential measurements, the system was able to reach a comparable performance level by the end of the traverse, and that the RNG extracted at each step tended to remain more constrained than that seen in the Monte-Carlo results.

\section*{Acknowledgments}

The research presented in this paper has been supported by National Science Foundation grant IIS-1320490 and a fellowship from the NASA Graduate Student Research Program. This work was performed by Cornell University and by the Jet Propulsion Laboratory, California Institute of Technology, under contract with the National Aeronautics and Space Administration.

\bibliographystyle{unsrtnat}
\bibliography{references}

\end{document}